# Research on a Camera Position Measurement Method based on a Parallel Perspective Error Transfer Model

Ning Hu,[1,*] Shuai Li,[2] and Jindong Tan[3]

[1]Ph.D. Candidate, Department of Mechanical, Aerospace and Biomedical Engineering, University of Tennessee, Knoxville, TN 37916, USA; email: nhu2@vols.utk.edu ; ning.hu.666@gmail.com (Corresponding author)

[2]Associate Professor, Department of Environmental Engineering Sciences, University of Florida, Gainesville, FL 32611, USA; email: shuai.li@ufl.edu

[3]Professor, Department of Mechanical, Aerospace and Biomedical Engineering, University of Tennessee, Knoxville, TN 37916, USA; email: tan@utk.edu

**Abstract**

Camera pose estimation from sparse correspondences is a fundamental problem in geometric computer vision and remains particularly challenging in near-field scenarios, where strong perspective effects and heterogeneous measurement noise can significantly degrade the stability of analytic PnP solutions. In this paper, we present a geometric error propagation framework for camera pose estimation based on a parallel perspective approximation. By explicitly modeling how image measurement errors propagate through perspective geometry, we derive an error transfer model that characterizes the relationship between feature point distribution, camera depth, and pose estimation uncertainty. Building on this analysis, we develop a pose estimation method that leverages parallel perspective initialization and error-aware weighting within a Gauss–Newton optimization scheme, leading to improved robustness in proximity operations. Extensive experiments on both synthetic data and real-world images, covering diverse conditions such as strong illumination, surgical lighting, and underwater low-light environments, demonstrate that the proposed approach achieves accuracy and robustness comparable to state-of-the-art analytic and iterative PnP methods, while maintaining high computational efficiency. These results highlight the importance of explicit geometric error modeling for reliable camera pose estimation in challenging near-field settings.

# 1. INTRODUCTION

With the rapid advancement of intelligent systems, **proximity operations**, characterized by transitions from far-field perception to near-field interaction, have become increasingly critical in a wide range of application domains, including space exploration, underwater missions, and surgical robotics [1]–[3]. Typical tasks such as on-orbit refueling, component replacement, spacecraft docking [4], unmanned aerial vehicle (UAV) landing [5]–[7], underwater vehicle docking [2], and autonomous surgical procedures [8] all require **highly reliable pose estimation under strong geometric constraints and adverse sensing conditions**. In such near-field scenarios, small perception errors may be systematically amplified through perspective geometry, leading to pronounced instability in downstream pose estimation and control.

Camera-based pose measurement remains one of the most widely adopted sensing modalities for proximity operations due to its noncontact nature, high spatial resolution, low cost, and immunity to electromagnetic interference. By extracting visual features from image observations and estimating the relative 3D pose between the camera and a target, vision-based methods have been successfully applied to helmet-mounted displays [6], augmented reality [2], [7], [8], and simultaneous localization and mapping (SLAM). However, **the reliability of camera-based pose estimation in near-field regimes is fundamentally limited by geometric sensitivity to measurement noise**, particularly when targets occupy a large portion of the field of view or exhibit nonuniform feature distributions.

At the core of camera pose estimation lies the well-known **perspective-n-point (PnP) problem** [9], which seeks to recover camera pose from a set of 3D–2D correspondences. Existing PnP solutions can be broadly categorized into **iterative** and **analytical** methods. Iterative approaches formulate pose estimation as a nonlinear optimization problem and typically achieve high numerical accuracy when sufficient features and good initialization are available. However, they are prone to local minima and incur significant computational cost, making them unsuitable for real-time or resource-constrained applications. Representative examples include the POSIT algorithm [10] and the OI algorithm [11].

Analytical PnP methods, in contrast, offer closed-form or near-closed-form solutions with favorable computational efficiency. The EPnP algorithm proposed by Lepetit *et al.* [12] was the first to achieve linear complexity with respect to the number of feature points. Subsequent extensions, such as RPnP [13] and OPnP [14], further improved numerical accuracy by exploiting polynomial formulations and global optimization strategies. Despite these advances, **analytical PnP methods remain highly sensitive to image noise and initialization quality**, particularly in near-field settings where perspective effects are strong. In practice, methods such as OPnP incur substantial computational overhead, limiting their applicability in scenarios where both robustness and efficiency are required, such as space, underwater, and surgical environments.

Most existing studies on camera pose estimation focus on **idealized sensing assumptions**, including uniform feature extraction accuracy and weak perspective approximations [15], [16]. Related efforts span feature extraction techniques [17]–[19], pose-solving algorithms [20]–[23], and error analysis methods [24]. However, **real-world near-field applications often violate these assumptions**. Factors such as solar radiation in space, strong illumination in operating rooms, and low-light conditions underwater introduce **heterogeneous, geometry-dependent measurement errors** that are not adequately captured by conventional PnP formulations. As a result, analytical methods that assume homogeneous noise distributions may exhibit systematic bias and instability, even when sufficient feature correspondences are available.

A key observation motivating this work is that **near-field PnP instability is not merely an algorithmic issue, but a geometric one**. In particular, the interaction between perspective projection, feature spatial distribution, and depth-dependent measurement uncertainty gives rise to **structured error propagation mechanisms** that directly influence pose estimation accuracy. Understanding and modeling this error transfer behavior is therefore essential for designing robust pose estimation algorithms that operate reliably across a wide range of near-field conditions.

To address this challenge, we present a **geometric error propagation framework for camera pose estimation based on a parallel perspective approximation** within the EPnP formulation. Rather than treating measurement noise as independent and identically distributed, the proposed approach explicitly models how image-space errors propagate through perspective geometry as a function of feature layout, camera depth, and viewing configuration. Building on this analysis, we introduce a pose estimation method that leverages parallel perspective initialization together with **error-aware weighting in a Gauss–Newton optimization scheme**, yielding improved robustness in proximity operations. In addition, guided by the derived error transfer model, we investigate the influence of feature point layout on pose stability and design a marker configuration tailored for near-field scenarios.

The proposed method is validated through extensive experiments on both simulated data and real-world images under diverse conditions, including strong illumination, surgical lighting, and underwater low-light environments. The results demonstrate that the proposed approach achieves accuracy and robustness comparable to state-of-the-art analytical and iterative PnP methods, while maintaining high computational efficiency across a wide range of near-field configurations.

The main contributions of this paper are summarized as follows:

1. **1. A geometric error propagation framework is developed to analyze depth- and layout-dependent instability in near-field PnP problems under perspective projection.**
2. **2. Based on the proposed error transfer analysis, the influence of feature spatial configuration on pose estimation robustness is systematically investigated, providing practical guidelines for feature layout design.**
3. **3. A parallel perspective–based pose estimation method with error-aware optimization is introduced to demonstrate how geometry-consistent modeling improves robustness and efficiency in near-field scenarios.**

The remainder of this paper is organized as follows. Section II reviews the EPnP algorithm and classical error modeling approaches for planar feature distributions. Section III presents the proposed parallel perspective–based pose estimation method and the associated error transfer model. Section IV reports comprehensive experimental evaluations on simulated and real-world datasets. Section V discusses the implications of the results and concludes the paper.

# 2. RELATED WORK

A single-image camera position measurement task involves estimating the position and attitude of a camera based on a set of spatial points and their positional information in an image. As shown in Fig. 1, the coordinates of a point in space $P_i^W$ in the target coordinate system $\left\{O^W - X^W Y^W Z^W\right\}$ are denoted as $P_i^W$, the coordinates of a point in the camera coordinate system $\left\{O^C - X^C Y^C Z^C\right\}$ are $P_i^C$, and the pixel coordinates in the image plane coordinate system $\left\{O^f - uv\right\}$ are $p_i\left(u_i, v_i\right)$. Once the camera has been calibrated, each parameter of the internal parameter matrix of the camera can be obtained; these parameters include the focal length f, the optical center coordinates $\left(u_0, v_0\right)$, and the physical length of the unit pixel $\left(d_x, d_y\right)$, such that $f_x = f / d_x$ and $f_y = f / d_y$. The resulting matrix is defined as follows:

$$K = \begin{bmatrix} f_x & 0 & u_0 \\ 0 & f_y & v_0 \\ 0 & 0 & 1 \end{bmatrix} \quad (1)$$

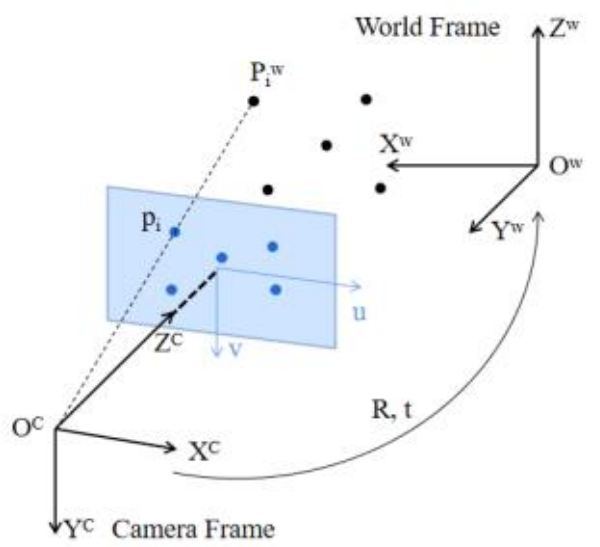

**Fig. 1.** Schematic of the perspective projection model for a camera.

According to the perspective projection model of the camera, the coordinates $P_i^C$ of a spatial point in the camera coordinate system are related to the projection coordinates $p_i\left(u_i, v_i\right)$ of the image plane as shown below:

$$\lambda_i \begin{bmatrix} u_i \\ v_i \\ 1 \end{bmatrix} = KP_c^i, i = 1, 2, ......n \qquad (2)$$

where $\lambda_i$ denotes the depth factor and $K$ denotes the internal reference matrix of the camera.

The coordinates of the flag point $P_i$ in the camera coordinate system $P_i^C$ and the corresponding coordinates in the world coordinate system $P_i^w$ satisfy the following relationship:

$$P_i^c = RP_i^w + t, i = 1, 2, ..., n \qquad (3)$$

where $R \in SO\left(3\right)$ denotes the rotation relationship between the world coordinate system and the camera coordinate system and $t \in R^3$ denotes the translation relationship between these two coordinate systems. Therefore, by using the above relationship model, the relative positions $R$ and $t$ between the camera coordinate system and the world coordinate system can be calculated by using n pairs of matched image coordinates $\left\{P_i, i = 1, 2, ...., n\right\}$ and their corresponding coordinates in the world coordinate system $\left\{P_i^W, i = 1, 2, ...., n\right\}$, while the internal reference of the camera $K$ is known.

### *2.1. The Classic EPnP Analytic Solution Algorithm and its Improvements*

The EPnP algorithm is a typical representative analytic PnP solution for the highly precise and fast position estimation algorithm proposed by Lepetit *et al*. [12] in 2009. This solution requires no iterative solving, has a temporal complexity of O(n), and has high computational accuracy; thus, it is presently considered one of the most efficient algorithms for estimating the position of a camera [25]. However, as an analytic algorithm, the EPnP algorithm is not robust to image noise and requires the use of complex equation-solving techniques during the parameter solving process, so the algorithm is not stable when fewer marker points are available.

The basic idea of EPnP is to represent n 3D points in the world coordinate system as a linear combination of 4 control points and then solve for the coordinates of the control points in the camera coordinate system according to the positional relationship between the spatial points and the control points, thereby obtaining the positional attitude of the camera by solving the absolute orientation problem. A point $P_i^w$ in the world coordinate system can be expressed as a linear combination of 4 noncoplanar virtual control points $C_j^w = [x_j^w, y_j^w, z_j^w]^T$ :

$$P_i^w = \sum_{j=1}^{4} \alpha_{ij} C_j^W \qquad (4)$$

where $\alpha_{ij}$ is the homogeneous barycentric coordinate of feature point $P_i^w$, which satisfies $\sum_{j=1}^{4} \alpha_{ij} = 1$, and the theoretically noncoplanar virtual control point $C_j^W$ can be chosen arbitrarily. In particular, the relationship between the homogeneous barycentric coordinates of the same sign point $P_i$ and the control point remains unchanged in different coordinate systems; i.e., the relationship between the homogeneous barycentric coordinates of the sign point $P_i$ and the control point $C_j^C = [x_j^C, y_j^C, z_j^C]^T$ in the camera coordinate system is also satisfied:

$$P_i^C = \sum_{j=1}^{4} \alpha_{ij} C_j^C \qquad (5)$$

Thus, the perspective projection model of the camera can be modified to the following:

$$\lambda_i \begin{bmatrix} u_i \\ v_i \\ 1 \end{bmatrix} = \begin{bmatrix} f_x & 0 & u_0 \\ 0 & f_y & v_0 \\ 0 & 0 & 1 \end{bmatrix} \sum_{j=1}^{4} \alpha_{ij} \begin{bmatrix} x_j^C \\ y_j^C \\ z_j^C \end{bmatrix} \quad (6)$$

Organizing this equation yields a system of equations with respect to $\left\{\left(x_j^C, y_j^C, z_j^C\right)\right\}_{J=1,\ldots,4}$:

$$\begin{cases} \sum_{j=1}^{4} \alpha_{ij} f_u x_j^c + \alpha_{ij}(u_c - u_i) z_j^c = 0 \\ \sum_{j=1}^{4} \alpha_{ij} f_v y_j^c + \alpha_{ij}(v_c - v_i) z_j^c = 0 \end{cases} \quad (7)$$

For n sign points, we obtain 2n equations:

$$\mathbf{M}_{2n\times 12}\mathbf{x}_{12\times 1} = 0 \quad (8)$$

The equation $x = [C_1^{cT}, C_2^{cT}, C_3^{cT}, C_4^{cT}]^T$. The solution of the above system of equations $\mathbf{x}$ lies in the zero space of the coefficient matrix $M$ and can be expressed as shown below:

$$\mathbf{x} = \sum_{i=1}^{N} \beta_i \mathbf{v}_i \quad (9)$$

where $\mathbf{v}_i$ is the eigenvector corresponding to the zero eigenvalue of $M^T M$. When the coefficient $\beta_i$ is determined, the coordinates of the four virtual control points under the camera coordinate system $C_j^C$ are also determined so that the coordinates of the marker points under the camera coordinate system $P_i^C$ can be calculated, and the position of the camera can be calculated according to the solution of the absolute orientation problem with the coordinates of the marker points in the world coordinate system and the coordinates of the camera coordinate system.

After the value of the coefficient $\beta_i$ is solved, the accuracy of the solution can be further improved via Gauss–Newton optimization, which aims to reduce the distance difference between the control points under two coordinate systems, namely, the camera coordinate system and the world coordinate system, with the following objective function:

$$\beta = \underset{\beta}{\operatorname{argmin}} \sum_{i,j=1,i<j}^{4} \| \mathbf{C}_i^c - \mathbf{C}_j^c \|^2 - \| \mathbf{C}_i^w - \mathbf{C}_j^w \|^2 \quad (10)$$

The EPnP algorithm has several shortcomings. Better results can be achieved when the number of 3D-2D points $n > 6$; however, the EPnP+GN algorithm performs erratically when the number of 3D-2D points is small, such as when $n = 4$ or 5. Further studies have revealed that the factor causing the instability of the algorithm is that the initial values of the coefficients $\beta_i$ obtained by the EPnP algorithm sometimes deviate severely from the correct values, which results in the subsequent Gaussian Newton optimization process failing to converge correctly. To solve the instability problem related to the computational results of the EPnP+GN algorithm, Chen [26] proposed the IEPnP algorithm, which simplifies the computational process of the EPnP+GN algorithm by using the weak perspective projection instead of the EPnP algorithm to estimate the initial $\beta_0$ value and then uses a Gauss–Newton optimization process identical to that of EPnP to improve the accuracy of the computational results. In practical use cases, however, the IEPnP algorithm is still not sufficiently stable when the marker point is closer to the camera, which is caused by the inaccuracy of the weak perspective projection estimate obtained when the marker point is closer.

Therefore, improving the accuracy of the initial value of the coefficients $\beta_i$, especially the accuracy of the initial values of the coefficients $\beta_i$ when the target is near the camera, is the key to improving the accuracy of the analytical PnP algorithm. In this paper, parallel perspective projection is introduced instead of EPnP to estimate the initial $\beta_0$, and then the same Gauss–Newton optimization process as that employed by EPnP is used to improve the accuracy of the calculation results. The parallel perspective projection is closer to the position of the true value than the weak perspective projection is when the marker point is closer to the camera. Therefore, the algorithm has greater stability.

### *2.2. Camera Attitude Measurement Error Modeling*

The analytical PnP method essentially solves the least-squares problem based on the positional parameters of the observed feature points, so the accuracy and stability of this solution algorithm are related not only to the extraction accuracy attained for the feature points but also to the layout and attitude of the marker points. Qu and Hou [27] took the P4P problem as the research object and used the spatial geometry method and error propagation theory to derive the analytical equation of the error function of the relevant estimation parameters and test variables in a monocular vision measurement scenario [27] (11) (12) (13); their work revealed the error law that affects the accuracy of attitude measurements, as shown in Fig. 2.

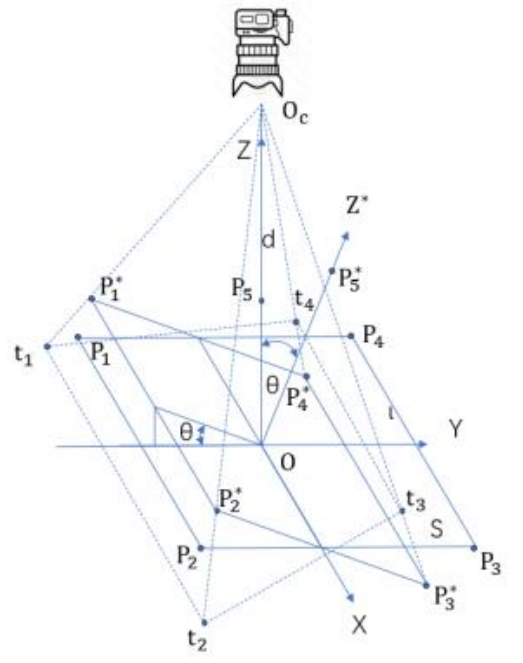

**Fig. 2.** Geometry of the pitch attitude solution.

$$\frac{\sigma_\gamma}{\sigma_p} \approx \frac{a}{2f}\left(\frac{d}{l}\right) \tag{11}$$

$$\frac{\sigma_\theta}{\sigma_p} \approx \frac{a}{2f}\left(\frac{d}{l}\right)^2 cos\theta \tag{12}$$

$$\frac{\sigma_\phi}{\sigma_p} \approx \frac{a}{2f}\left(\frac{d}{l}\right)^2 \cos\phi \tag{13}$$

where the target feature point is located in the plane $S$ in the camera coordinate system $\{O-X-Y-Z\}$; the azimuth, pitch, and tilt angles are $\gamma,\theta,\phi$, respectively; $a$ is the camera cell size; $f$ denotes the focal length of the camera; $d$ is the distance from the plane $S$ in the coordinate system to the camera; $l$ signifies the distance of the neighboring target point; $P$ represents the target feature point; $P^*$ is the position of point $P$ rotated about the $X$ -axis; $t$ denotes the projection of point $P^*$ onto the S-plane point; and $\sigma_p$ is the feature point-based image extraction error for determining the estimation errors of the pitch angle and the tilt angle. $\sigma_\phi$ is the feature point image extraction error, $\sigma_\gamma$ is the azimuth angle estimation error, $\sigma_\theta$ is the pitch angle estimation error and is the tilt angle estimation error.

The error model shows that when the monocular vision measurement parameters are determined, the azimuth angle measurement error is independent of the azimuth angle value, is proportional only to $\frac{d}{l}$, and is positively correlated with the feature point-based image position extraction accuracy. The estimation errors related to the pitch angle and tilt angle are positively correlated with the cosine values of their own angles, positively correlated with the feature point-based image position extraction accuracy, and proportional to the square of $\frac{d}{l}$.

Qu and Hou [27] verified the correctness of the above mathematical model of the camera position estimation error based on plane-distributed feature points and the validity of the step-by-step error analysis method based on error propagation theory through experimental simulations. The error model reveals that the four feature points are distributed planarly, and although the unique target solution can be obtained, the estimation errors of the pitch and tilt angles are most affected by the feature points extraction accuracy when the plane in which the feature points are located is perpendicular to the optical axis of the camera. Therefore, when approaching the target to implement high-precision near-field operations, the marker point plane must be distributed at an angle to the optical axis of the camera, or a nonplanar distribution must be used. In a near-field operation task, the camera will be sufficiently close to the target in the final operation stage, and the field of view of the camera will be restricted, which will affect the distribution position of the feature points. Therefore, the three-dimensional feature point layout and the error step-by-step analysis method based on error propagation theory are adopted in this paper to study methods for improving the accuracy and robustness of PnP attitude estimation.

Unlike existing uncertainty-aware or globally optimal PnP solvers that primarily focus on algebraic optimization, this work emphasizes geometric error transfer mechanisms that govern near-field pose instability.

# 3. Camera Position Estimation Algorithm Based on Parallel Perspective Projection Error Modeling

*3.1. Optimization Idea of the EPnP Algorithm based on the Parallel Perspective Projection Model*

In this paper, across the different stages of near-field operations, the distance between the target and camera gradually decreases; thus, in combination with the problems posed by the EPnP+GN and IEPnP algorithms, a new weighted optimization algorithm based on EPnP is proposed. First, the parallel perspective projection model is used to obtain initial estimates of the positional attitude of the target $R_0$ and $t_0$ according to the relation shown in (14):

$$C_j^c = R_0 C_j^w + t_0 \qquad (14)$$

The coordinates of the control point in the camera coordinate system can be obtained from $C_j^c$, which in turn can be obtained from $x = [C_1^{cT}, C_2^{cT}, C_3^{cT}, C_4^{cT}]^T$ as follows:

$$\mathbf{x} = \begin{bmatrix} \mathbf{v}_1 & \mathbf{v}_2 & \mathbf{v}_3 & \mathbf{v}_4 \end{bmatrix} \begin{bmatrix} \beta_1 \\ \beta_2 \\ \beta_3 \\ \beta_4 \end{bmatrix} = \mathbf{V\beta} \qquad (15)$$

Thus, the initial values of the coefficients $\beta = [\beta_1, \beta_2, \beta_3, \beta_4]^T$ can be obtained when $V$ and $\mathrm{x}$ are known:

$$\beta_0 = (V^T V)^{-1} V^T \mathrm{x} \qquad (16)$$

Afterward, the accuracy of the coefficients $\beta$ is improved by introducing the factors that influence the accuracy of attitude estimation through the parallel perspective error model in the G-N optimization function.

*3.2. Estimation of the Initial Attitude Value via Parallel Perspective Modeling*

As shown in Fig. 3, the parallel perspective projection model was proposed by Horaud *et al*. [28] for approximating pose solutions under the perspective projection model via an iterative parallel perspective solving process. The parallel perspective projection model can be regarded as a first-order approximation of the perspective projection, and the parallel perspective projection equations can be expressed as follows by setting the camera positions $R = [i, j, k]^T$ and $t = [t_x, t_y, t_z]^T$:

$$\begin{cases} u_i = \dfrac{\mathbf{i}^T - u_0 \mathbf{k}^T}{t_z} \cdot \mathbf{P}_i^w + u_0 \\ v_i = \dfrac{\mathbf{j}^T - v_0 \mathbf{k}^T}{t_z} \cdot \mathbf{P}_i^w + v_0 \end{cases} \qquad (17)$$

where $u_0 = t_x / t_z$ and $v_0 = t_y / t_z$. $t_z$ is the offset of the origin of the camera coordinate system in the Z-axis of the target coordinate system, which represents the distance from the camera to the target in the direction of the optical axis of the camera during camera motion. Therefore, the value is defined as the depth of the camera. Definition:

$$\begin{cases} \mathbf{I}_p = \dfrac{\mathbf{i}^T - u_0 \mathbf{k}^T}{t_z} \\ \mathbf{J}_p = \dfrac{\mathbf{j}^T - v_0 \mathbf{k}^T}{t_z} \end{cases} \qquad (18)$$

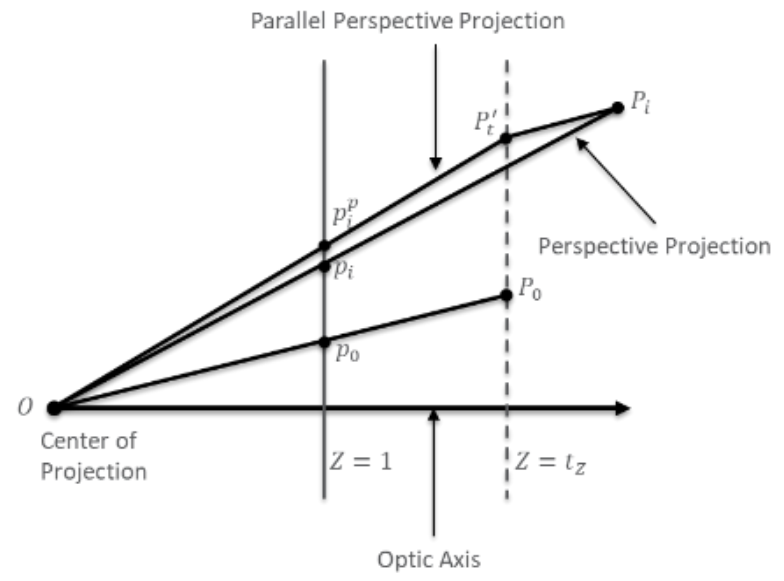

**Fig. 3.** Schematic of parallel perspective projection. $P_0$ is the origin of the target coordinate system, and $P_i$ represents any point in the target coordinate system. $p_i$ and $p_0$ are the projections of the unit phase planes $P_i$ and $P_0$, respectively.

If the spatial coordinates and image coordinates of the four feature points are known, then $\mathbf{I}_p$ and $\mathbf{J}_p$ can be solved linearly via (17) such that taking the modulus of $\mathbf{I}_p$ and $\mathbf{J}_p$ yields two solutions for $t_z$. By taking the average of the two solutions as the estimate of $t_z$, the three elements of the translation vector can be obtained as follows.

$$\begin{cases} t_z = \frac{1}{2}\left( \frac{\sqrt{1+u_0^2}}{\| \mathrm{I}_p \|} + \frac{\sqrt{1+v_0^2}}{\| \mathrm{I}_p \|} \right) \\ t_x = u_0 t_z \\ t_y = v_0 t_z \end{cases} \tag{19}$$

According to R, the orthogonality property is as follows:

$$\begin{aligned} \mathbf{k} &= \mathbf{i} \times \mathbf{j} \\ &= t_z^2 \mathbf{I}_p^T \times \mathbf{J}_p^T - t_z u_0 \mathbf{J}_p^T \times \mathbf{k} + t_z v_0 \mathbf{I}_p^T \times \mathbf{k} \end{aligned} \tag{20}$$

If we let $[\cdot]_\times$ denote the antisymmetric matrix corresponding to the 3-dimensional vector, the above equation becomes:

$$\mathbf{k} = \left( \mathbf{I} - t_z v_0 [\mathbf{I}_p]_\times + t_z u_0 [\mathbf{J}_p]_\times \right)^{-1} t_z^2 (\mathbf{I}_p \times \mathrm{J}_p) \tag{21}$$

Substituting these results into (18) yields the corresponding $\mathbf{i}$ and $\mathbf{j}$.

*3.3. The Line Segment Error Transfer Model*

From (17),

$$\begin{cases} t_z u_i = \mathbf{i}^T \cdot \mathbf{P}_i^w + t_z u_0 - u_0 \mathbf{k}^T \cdot \mathbf{P}_i^w \\ t_z v_i = \mathbf{i}^T \cdot \mathbf{P}_i^w + t_z v_0 - v_0 \mathbf{k}^T \cdot \mathbf{P}_i^w \end{cases} \tag{22}$$

The above equation can be organized as follows:

$$\begin{cases} t_z (u_i - u_0) = P_{ix}^c - t_x - u_0 (P_{iz}^c - t_z) \\ t_z (v_i - v_0) = P_{iy}^c - t_y - v_0 (P_{iz}^c - t_z) \end{cases} \tag{23}$$

where $P_{ix}^c$, $P_{iy}^c$, and $P_{iz}^c$ denote the x-, y-, and z-axis coordinates of marker $P_i$ in the camera coordinate system, respectively.

$$\begin{cases} d_{P_{ix}} = P_{ix}^c - t_x \\ d_{P_{iy}} = P_{iy}^c - t_y \\ d_{P_{iz}} = P_{iz}^c - t_z \\ u_{ix} = u_i - u_0 \\ v_{ix} = v_i - v_0 \end{cases} \tag{24}$$

Then, (23) can be converted to the following:

$$\begin{cases} t_z u_{ix} = d_{P_{ix}} - u_0 d_{P_{iz}} \\ t_z v_{ix} = d_{P_{iy}} - v_0 d_{P_{iz}} \end{cases} \tag{25}$$

The point $P_0$ is defined as the origin of the target coordinate system, the distance from the spatial point $P_i$ to the origin of the target coordinate system $P_0$ is $d_{i-o}$, the projection length of $P_0P_i$ in the normalized image plane is $m_{i-o} = \sqrt{u_{ix}^2 + v_{ix}^2}$, the distance from the origin of the world coordinate system to the optical axis of the camera is the projection length of $m_o = \sqrt{u_0{}^2 + v_0{}^2}$, and the angle between $P_0P_i$ and the optical axis of the camera is set as $\beta$ , which is obtained from (25):

$$(d_{i-o}\sin\beta)^2 = d_{P_{ix}}^2 + d_{P_{iy}}^2 = t_z^2 m_{i-o}^2 + m_o^2 d_{P_{iz}}^2 + 2(u_{ix}u_0 + v_{ix}v_0)t_z d_{P_{iz}} \tag{26}$$

This can be obtained from the principle of error propagation:

$$4d_{i-o}^{2X2-2}(sin\beta)^{2X2}\delta_{d_{i-o}}^2 + 4d_{i-o}^{2X2}sin\beta^{2X2-2}cos\beta\delta_\beta^2 = \delta_{m_{i-o}}^2(2t_z^2 m_{i-o})^2 + \delta_{m_o}^2\left(2d_{P_{i-o}}^2 m_o\right)^2 + \delta_{u_{ix}}^2\left(2u_o t_z d_{P_{ix}}\right)^2 + \delta_{v_{ix}}^2\left(2v_o t_z d_{P_{ix}}\right)^2 \tag{27}$$

The distance between two points on the line segment $P_0P_i d_{i-o}$ is defined as an indirect measurement, the projection of the line segment $P_0P_i$ is defined in the image $m_{i-o}$, and the projection of the plumb line from the origin of the target coordinate system $P_0$ to the optical axis of the camera in the image $m_o$ is defined as a direct measurement; then, according to the above equation, we have the following:

$$\begin{cases} \dfrac{\delta_{d_{i-o}}}{\delta_{m_{i-o}}} = \dfrac{2t_z^2 m_{i-o}}{2d_{i-o}(\sin\beta)^2} = \dfrac{t_z^2 m_{i-o}}{d_{i-o}(\sin\beta)^2} \\ \dfrac{\delta_{d_{i-o}}}{\delta_{m_o}} = \dfrac{2d_{P_{iz}}^2 m_o}{2d_{i-o}(\sin\beta)^2} = \dfrac{d_{P_{iz}}^2 m_o}{d_{i-o}(\sin\beta)^2} \\ \dfrac{\delta_{d_{i-o}}}{\delta_\beta} = \dfrac{2d_{i-o}^2\sin\beta(\cos\beta)^{0.5}}{2d_{i-o}(\sin\beta)^2} = \dfrac{d_{i-o}(\cos\beta)^{0.5}}{\sin\beta} \end{cases} \tag{28}$$

The above equation shows that the quotient of the measurement error of the indirect measurement value $d_{i-o}$ and the measurement error of the direct measurement value $m_{i-o}$ is proportional to the square of the camera depth, proportional to the magnitude of the direct measurement value $m_{i-o}$, inversely proportional to the length of the line segment, and inversely proportional to the square of the sine of the angle of the line segment with the optical axis $\beta$ . Since $P_{iz}^c$ is the z-axis coordinate of the feature point in the camera coordinate system, it is close to the z-coordinate of the origin of the target coordinate system $t_z$, and this distance is much larger than the length of the line segment. Therefore, it can be assumed that the direct measurement value $m_o$ has a significant influence on the measurement error of the line segment.

The angle between the line segment and the optical axis of the camera $\beta$ is defined as an indirect measurement, and the projection of the line segment in the image $m_{i-o}$ and the projection of the plumb line from the origin of the target coordinate system $P_0$ to the optical axis of the camera in the image $m_o$ are defined as direct measurements. Then, according to the above equation, we have the following:

$$\begin{cases} \dfrac{\delta_\beta}{\delta_{m_{i-o}}} = \dfrac{2t_z^2 m_{i-o}}{2d_{i-o}^2\sin\beta(\cos\beta)^{0.5}} = \dfrac{t_z^2 m_{i-o}}{d_{i-o}^2\sin\beta(\cos\beta)^{0.5}} \\ \dfrac{\delta_\beta}{\delta_{m_o}} = \dfrac{2d_{P_{iz}}^2 m_o}{2d_{i-o}^2\sin\beta(\cos\beta)^{0.5}} = \dfrac{d_{P_{iz}}^2 m_o}{d_{i-o}^2\sin\beta(\cos\beta)^{0.5}} \end{cases} \tag{29}$$

The above equation shows that the quotient of the error of the indirect measurement $\beta$ and the measurement error of the direct measurement $m_{i-o}$ is proportional to the square of the depth of the camera, proportional to the magnitude of the direct measurement $m_{i-o}$, inversely proportional to the length of the line segment $d_{i-o}$ squared, and inversely proportional to the product of the square root of the sine of the angle $\beta$ and the cosine of the angle $\beta$ . Since $P_{iz}^c$ is the z-axis coordinate of the feature point in the camera coordinate system, it is close to the z-coordinate of the origin of the target coordinate system $t_z$, and this distance is much larger than the length of the line segment. Therefore, it can be assumed that the direct measurement $m_o$ has a significant effect on the error of the indirect measurement $\beta$ .

When the order of the coordinate system transformation is the ZXZ order of the Euler angles, if the point $P_0$ is the origin of the target coordinate system and the spatial point $P_i$ is on the y-axis of the target coordinate system, then the angle $\beta$ is the residual angle of the pitch angle of the target coordinate system. If the spatial point $P_i$ is located on the z-axis of the target coordinate system, then the angle $\beta$ is the pitch angle of the target coordinate system. In the planar feature point distribution scenario, the error transfer model in this study is basically consistent with the transfer model concerning the angle measurement error of the Qu Yuyi frequency [27] and covers the spatial feature point distribution scenario.

### 3.4. Optimization of the Gauss–Newton Objective Function

The objective of Gauss–Newton optimization is to reduce the distance between the control points in two coordinate systems: the camera coordinate system and the world coordinate system. Therefore, the Gauss–Newton objective function optimization scheme in the EPnP algorithm is given weights on the basis of the error transfer relationship via the error model between the points.

Regardless of whether point $\mathbf{P}_i$ is on the target coordinate system axis, the position error of the target coordinate system (the measurement error of the line segment passing through the origin of the target coordinate system) is proportional to the square of the camera depth, inversely proportional to the length of the line segment, and directly proportional to the values of the direct measurements $m_o$ and $m_{i-o}$. $d_{i-o}$ is the length of the line segment, which is much smaller than the camera depth (whether it is $d^2{}_{i-o}$ or $d_{i-o}$), and the value of $\sin\beta(\cos\beta)^{0.5}$ or $(\sin\beta)^2$ is less than 1. Therefore, either $d^2{}_{i-o}\sin\beta(\cos\beta)^{0.5}$ or $d_{i-o}(\sin\beta)^2$ can be optimized to $d^2{}_{i-o}$. The optimization weights are set to the following:

$$w_{ij} = \frac{d_{i-o}^2}{t_z^2 m_o m_{i-o}} \tag{30}$$

Thus, the Gauss–Newton objective function optimization equation becomes:

$$\beta = \underset{\beta}{\operatorname{argmin}} \sum_{i,j=1,i<j}^{4} w_{ij} \left| \| \mathbf{C}_i^c - \mathbf{C}_j^c \|^2 - \right| \| \mathbf{C}_i^w - \mathbf{C}_j^w \|^2 \tag{31}$$

### 3.5. Algorithmic Pseudocode

Input: 2D/3D point pair coordinates

Outputs: rotation coordinate matrix of the camera, translation matrix, bit position of the camera

Define four noncoplanar virtual control points in the 3D world coordinate system to establish the control point coordinate system;

Use (17), (18), (19), (20), and (21) of the perspective projection model to calculate the transformation matrix and the initial value of the translation matrix.

Calculate the coordinates of the control point under the camera coordinate system according to (9);

Calculate and the initial $\boldsymbol{\beta}_0$ value according to (14), (15), and (16);

According to (30) and (31), use the Gauss–Newton method to solve for the optimal solution $\boldsymbol{\beta}$;

Calculate the coordinates of the control points under the camera coordinate system according to (9), and calculate the coordinates of the target feature points under the camera coordinate system according to (5);

Based on the coordinates of the feature points in the image coordinate system and the target object coordinate system, calculate the position of the camera according to [29].

It is worth noting that the proposed error propagation analysis is derived under general perspective projection assumptions and does not rely on a specific marker geometry. While a biprism feature layout is adopted in this work for experimental validation, the theoretical conclusions regarding depth-dependent error amplification and layout-induced robustness apply to general noncoplanar feature configurations commonly encountered in near-field perception tasks.

# 4. EXPERIMENTS AND RESULTS

### 4.1. Experimental Setup and Platform Construction

#### 1) Experimental Setup

The accuracy of the camera position estimation process is affected by a variety of factors, including the number of feature points, the noise magnitude, the computing time [30], the quotient of the distance between the camera and the target and the size of the target, and the distance from the target to the optical axis of the camera [28]. To validate the proposed algorithm, the proposed algorithm is validated on simulated data and real image data and compared with other well-known algorithms, including an accurate orthogonal iterative method (LHM) [29], an efficient noniterative method (EPnP) [12], the suboptimal method (RPnP) [13], direct least-squares DLS [9], [31] with an accurate analytical solution (OPnP) [14], the parallel algorithm perspective [28] (which does not have a Gaussian Newton optimization part in this paper), and a typical linear method (DLT) [32], [33]. All algorithmic codes are publicly available on our website and are MATLAB implementations, and they are executed on 8-core personal computers with AMD Ryzen 73,700X CPUs and 32 GB of RAM, as shown in Fig. 4.

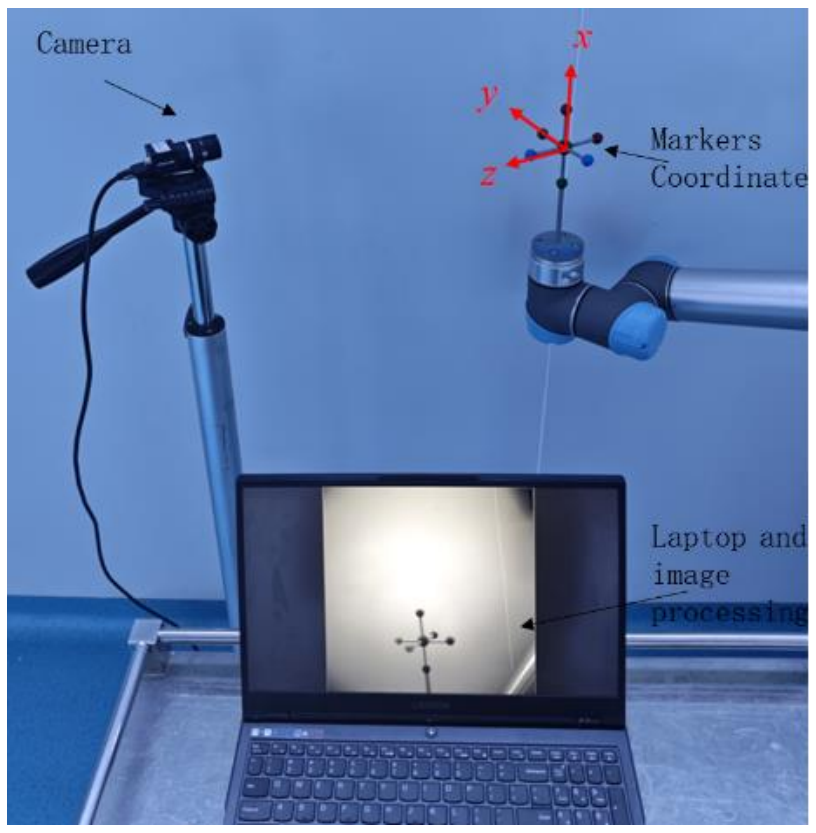

**Fig. 4.** Overview of the experimental platform.

### 2) *Experimental Platform*

In Fig. 5, the basic components of the camera pose measurement platform consist of a target image acquisition camera (a hikvision MVL-HF0628-05S camera with an F2.8 M12 interface lens possessing a 6-mm focal length), a coordinate system of feature points representing the target object, an image processing and pose estimation controller (an 8-core personal computer with an AMD Ryzen 73,700X CPU and 32 GB of RAM), and an image processing and pose estimation controller (an 8-core personal computer with an AMD Ryzen 73,700X CPU and 32 GB of RAM). To test the performance of the developed camera pose estimation algorithm, the experimental platform also includes a 6-joint robot (UR5) that changes the position and pose of the target object, along with a light source that simulates different ambient lights. To facilitate testing, the feature point coordinate system designed in this paper is mounted on the end of the robot, and the camera is fixed. The robot drives the feature point coordinate system to move, the camera collects an image of the feature point coordinate system after each movement, and finally, the image processing and attitude estimation controller completes the image acquisition and attitude estimation processes.

### 3) *Design of the Marking Points*

The target position information carried by the target feature points is directly related to the shape and layout of the target feature points. According to the error transfer model of parallel perspective projection, the measurement error of the azimuth angle is independent of the attitude of the coordinate system constructed by the target feature point and is proportional only to the quotient of the camera depth to the target size. The measurement errors induced for the pitch and tilt angles are proportional not only to the square of the quotient of the camera-target distance and the target size but also to the attitude of the coordinate system constructed by the target feature points. When the distribution of the target feature points is noncoplanar, the stereoscopic layout of feature points can reduce the attitude measurement error of the target coordinate system constructed by the feature points. Therefore, a stereo layout is chosen for the target feature points of the attitude measurement system studied in this paper is chosen.

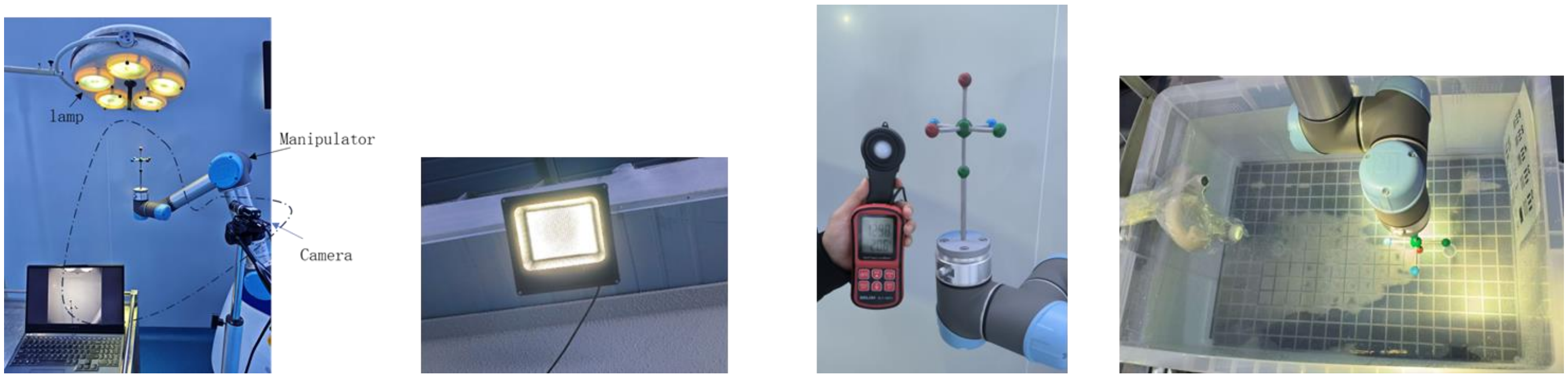

**Fig. 5.** (a) Atmospheric experimental platform structure for testing the proposed camera position measurement algorithm. (b) Test environment containing simulated strong light radiation with a light intensity level of 129.8 lux. (c) Simulated hand operating room with shadowless lamp irradiation-based test environment (light intensity=114.8 lux). (d) Underwater experimental platform structure for testing the developed camera position measurement algorithm.

According to the PnP perspective projection model, there are 12 unknown parameters, and for each 2D–3D feature point pair, 2 equations can be constructed, and 6 feature point pairs can determine a uniquely solved camera pose. However, when the number of target feature points exceeds 12, this setting has little effect on the accuracy of the obtained target pose measurement. Therefore, without loss of generality, the number of feature points employed for the experimental validation of the algorithm proposed in this paper adopts is 6, and the points are arranged in a biprism layout.

The common features of solid targets include points, lines, and curves. On the 2D image of a PnP problem, the positions and shapes of these features may change. According to the perspective transformation property, the intersection of straight lines and the center of a sphere in space exhibit perspective invariance; therefore, the center of the sphere is used as the feature point to design the 6-point biprism target used in this study (Fig. 6).

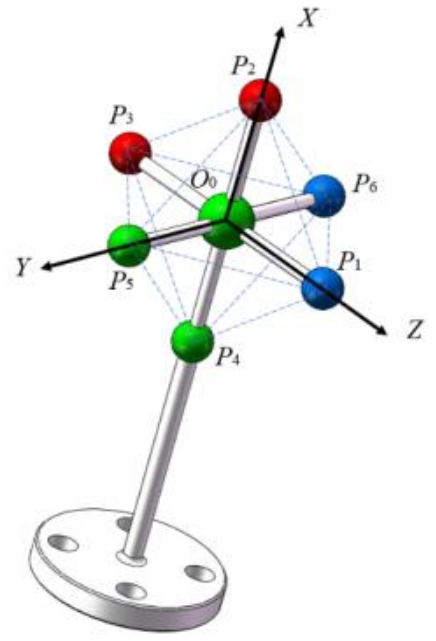

**Fig. 6.** Feature point structure in the target coordinate system.

The 6-point biprism target takes the 6 vertices of the biprism ($P_1$, $P_2$, $P_3$, $P_4$, $P_5$, and $P_6$) and the center of mass of the corresponding cone $O_0$ as the feature points of the attitude measurement algorithms and combines them with the geometric constraints between the feature points (all sides of the biprism are equal) to compare and analyze the characteristics of different attitude measurement algorithms. The dashed lines form the sides of the bipyramid; $P_1$, $P_2$, $P_3$, $P_4$, $P_5$, and $P_6$ are the vertices of the bipyramid; $O_0$ is the center of mass of the bipyramid; $P_1$,$P_2$,$P_3$,$P_4$, and $O_0$ are coplanar; and $P_2$, $P_4$, $P_5$, $P_6$, and $O_0$ are coplanar. The mounting base of the marker point coordinate system is connected to the robot end so that the x-axis of the marker point coordinate system coincides with the axis containing the robot end joints. $O_0$ is defined as the origin of the biprism coordinate system, $O_0P_1$ is the z-axis of the coordinate system, $O_0P_2$ is the x-axis of the coordinate system, and $O_0P_5$ is the y-axis of the coordinate system. To compare the EPnP algorithm and satisfy the need for 4 control points in the EPnP algorithm, by using the symmetry property of the bipyramidal body, the marker points at the origin of the coordinate axes and the 3 markers in the positive directions of the x, y and z-axes are directly employed as the control points for precision calculations, and the distance from the vertex of the bipyramidal cone to the center of mass is set to L . Then, $C_1^w = [L, 0,0]^T$ , and $C_2^w = [0, L, 0]^T C_3^w = [0,0, L]^T C_4^w = [0,0,0]^T$. To verify the effects of different object distances and marker point sizes on the algorithm, L is set to 50 mm.

*4.2. Experimental Procedure and Data Processing Strategy*

The data of this experiment include two parts: simulated data and real image data.

*1) Simulation Data Generation Methods*

After setting the internal reference matrix of the camera, according to the camera perspective projection model and (1), (2), and (3), the coordinates of the feature points in the target coordinate system, camera coordinate system, and image coordinate system are computed by using randomly generated spatial feature points, a rotation matrix, and a translation matrix to validate the algorithm developed in this study and the other comparison algorithms. Neglecting camera distortion, the internal reference matrix of the camera is set as follows:

$$cam_ = \begin{bmatrix} 1301.473508 & 0 & 653 \\ 0 & 1300.926193 & 508 \\ 0 & 0 & 1 \end{bmatrix}$$

Taking the focus of the camera as the reference, the feature points $P_i^c$ under the camera coordinate system are randomly distributed in a [-500,500] m×[-500,500] mm×[500,1500] mm rectangle. When the number of feature points $P_i^c$ is 6, the distribution of the feature points (including the control points required by this algorithm) is shown in the marker points presented in Fig. 4, and the origin of the coordinate system formed by the feature points $O_0$ is randomly selected within the range of [-500,500] m×[-500,500] mm×[500,1500] mm, with the focus of the camera used as a reference.

The Euler angles are randomly generated and then converted into a rotation matrix R. The translation vector T(Tx,Ty,Tz),Tz direction is defined as d (see the error transfer model), and the coordinates under the world coordinate system of the reference point and the coordinates of the image coordinate system are obtained through (1), (2), and (3). Gaussian noise is then added to the image coordinates (noise with a variance of 0.2 or noise with a variance of 0~5 pixels and a step size of 0.5 pixels). The coordinates under the image coordinate system and the coordinates under the world coordinate system with added noise are input into the algorithm developed in this study and the comparison algorithms to obtain the corresponding R and T matrices, and the rotation angle and the shift angle are calculated via (32) to obtain the angular error and the shift error.

$$R_{ZXZ}(\theta,\beta,\alpha) = \begin{bmatrix} -s\alpha c\beta s\theta + c\alpha c\theta & -s\alpha c\beta c\theta & s\alpha s\beta \\ c\alpha c\beta s\theta + s\alpha c\theta & c\alpha c\beta c\theta & -c\alpha s\beta \\ s\beta s\theta & s\beta s\theta & c\beta \end{bmatrix} \quad (32)$$

The transformation matrix $R_{ZXZ}$ is a matrix rotated around the Euler angles in the following order: the z-axis, new x-axis, and new z-axis.

Here, $\alpha$ is the rotation angle around the Z-axis of the coordinate system, $\beta$ is the new X-axis (N-axis) rotation angle obtained after the Z-axis is rotated, and $\theta$ is the new Z-axis rotation angle obtained after the N-axis is rotated.

*2) Experimental Procedures and Data based on Real Images*

Light sources with different intensities are used to simulate an atmospheric environment with strong light, the shadowless light environment of an operating room, and an underwater low-illumination environment. The robot of the experimental platform carries the coordinate system of the marker points to move in different environments and collects and retains the images of the target feature points on the image plane of the camera. The coordinates of the target feature points on the image are manually extracted via conventional methods such as the Hough transform and are retained as a txt file for use in subsequent research.

The camera resolution is set to 1280x720, and the distortion parameters are -0.087355, 0.204686, 0.000000, 0.000000, and 0.000000. The internal reference matrix of the camera is as follows:

$$cam_ = \begin{bmatrix} 1301.473508 & 0 & 653 \\ 0 & 1300.926193 & 508 \\ 0 & 0 & 1 \end{bmatrix}$$

The virtual light parameters are set as follows.

Atmospheric ambient light setting: light intensity = 129.8 lux.

Shadowless light setting of the operating room: light intensity = 114.8 lux.

Underwater low-light setting: light intensity = 100 lux.

The initial position of the 6-point feature point coordinate system is parallel to the camera coordinate system, with the origin on the main camera axis located 100 mm from the camera. Along the optical axis of the camera, the position of the feature point coordinate system is moved (50 mm from the center of the spherical feature point), starting from 200 mm away from the camera, with a step size of 200 mm, up to 1,000 mm; at each dwell position of the optical axis of the camera, in the plane perpendicular to the optical axis, the point moves along the X-direction by 50 mm and 100 mm. At each dwell position of the feature point coordinate system, the X-axis is rotated, and the feature point coordinate system is adjusted. At each dwell position of the feature point coordinate system, the X-axis is rotated, and the pitch angle of the feature point coordinate system is adjusted, starting from 0 degrees (with a step angle of 2 degrees), and rotated 90 degrees in the positive and negative directions, to collect the feature point image at each angular position. Since lateral movement causes the marker point to move out of the field of view when the origin is within 200 mm of the camera, only the photo of the marker point at the position of the optical axis point can be captured, so a total of 756 (1x13x18x3: 1 feature type, 6 distances + 8 panning positions, 18 angles, and 3 scenes) images are captured. The image coordinates of the marker points are extracted via the conventional marker point extraction algorithm and recorded in a txt file for subsequent research.

*3) Image Data Processing*

From the experimental data, we extract the pose data of the feature points in the target coordinate system and the camera coordinate system, as well as their coordinate positions in the image. We then use the algorithm studied in this paper and the comparison algorithms to estimate the pose and movement of the camera. Considering that the cumulative pose movement error has a large impact on the results, this paper adopts the relative error method to calculate the error of the pose measurement algorithm. For each image, the displacement matrix $t$ and rotation matrix $R_i$, $i = 1,2,\ldots,n$ corresponding to that image are calculated, the rotational pose $R_{i-1}$ of the previous image is taken as the base, and the relative relationship between the rotational pose of image $i$ and that of image $i-1$ ($R_{1,i}$) is calculated. Since the rotation angle of the coordinate system of the rotating feature point of the robotic arm is fixed, image $i$ is different from image i-1. The rotational attitude of the true relative relationship between these images $R_{1,i}^{true}$ is known; the $i$th image rotation angle error and displacement error are calculated as follows:

$$E_{rot}(degrees) = \max_{k\in\{1,2,3\}} cos^{-1}\left(\mathbf{r}_{k,true}^{T}\mathbf{r}_k\right)\times\frac{180}{\pi} \quad (33)$$

$$E_t(\%) = \| \mathbf{t}_{true} - \mathbf{t} \| / \| \mathbf{t}_{true} \|$$

where $r_{k,true}$ and $r_k$ are the $k$th columns of $R_{1,i}^{true}$ and $R_{1,i}$, respectively.

$t$ is the displacement matrix of the coordinate system of the camera relative to the target coordinate system, and $t_{true}$ is the

movement vector of the feature point coordinate system driven by the rotation of the robotic arm.

## *4.3. Experimental Results and Analysis*

### *1) Experimental Results and Analysis of Simulation Data*

- The influence of the number of feature points on the accuracy of the proposed algorithm

This experiment evaluates the influence of geometric constraint redundancy on pose estimation stability, as predicted by the proposed error propagation analysis, where increasing the number of noncoplanar feature points reduces the impact of individual measurement errors.

This experiment is intended to verify the effects of the number of feature points with 3D-2D point information on the accuracies of different position estimation algorithms. In the simulated data experiments, the number of input feature points ranges from 4 to 20, 200 cycles are chosen, white noise with a variance of 0.2 is added, and 80 camera focal lengths are chosen to verify the accuracy of the algorithms. The results of the experiments are shown in Figs. 7a and b, and through an analysis, we can draw the following conclusions.

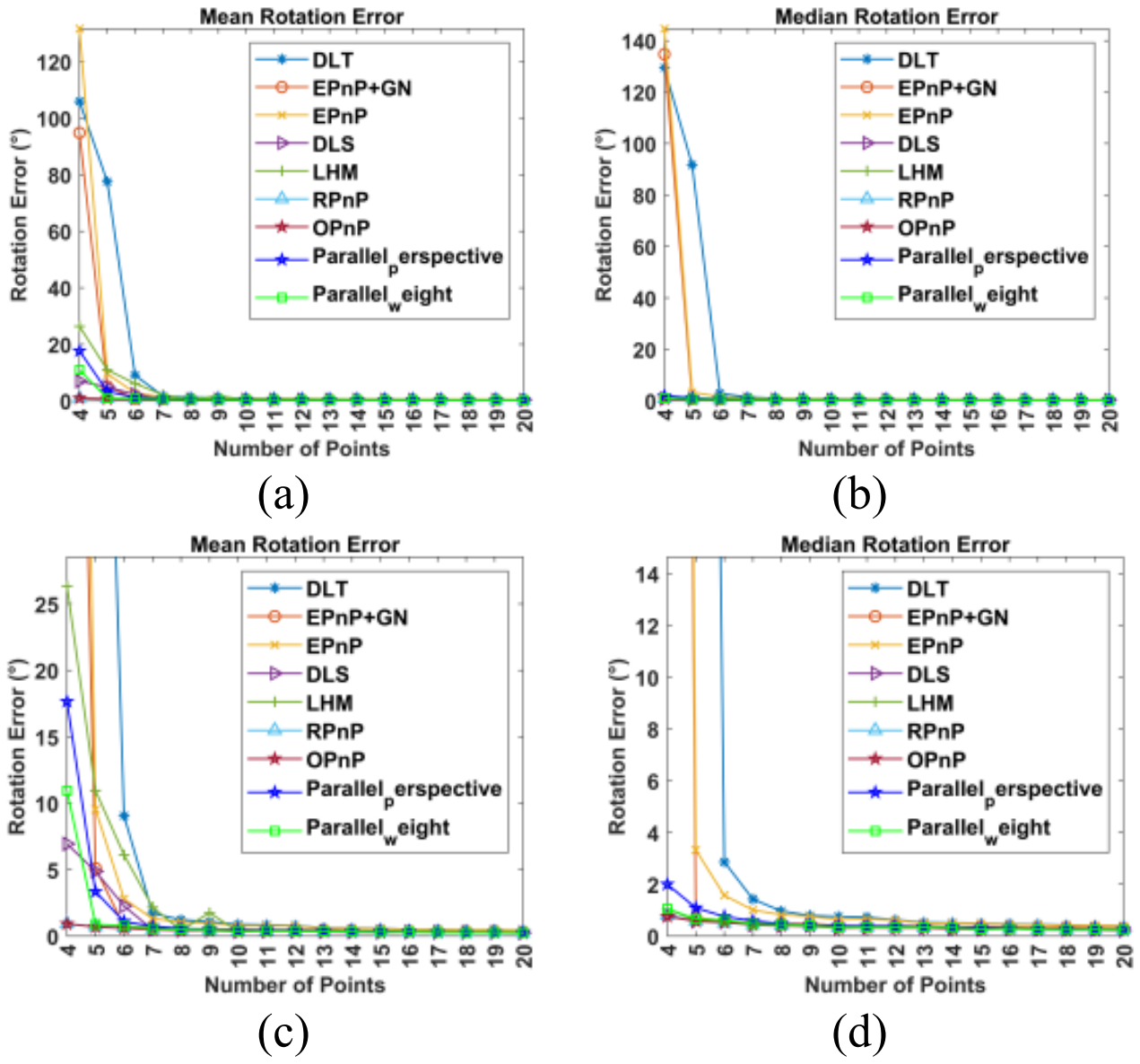

**Fig. 7a.** Effects of the number of feature points on the accuracies of attitude angle estimates ((a) and (b) are global plots, while (c) and (d) are local plots).

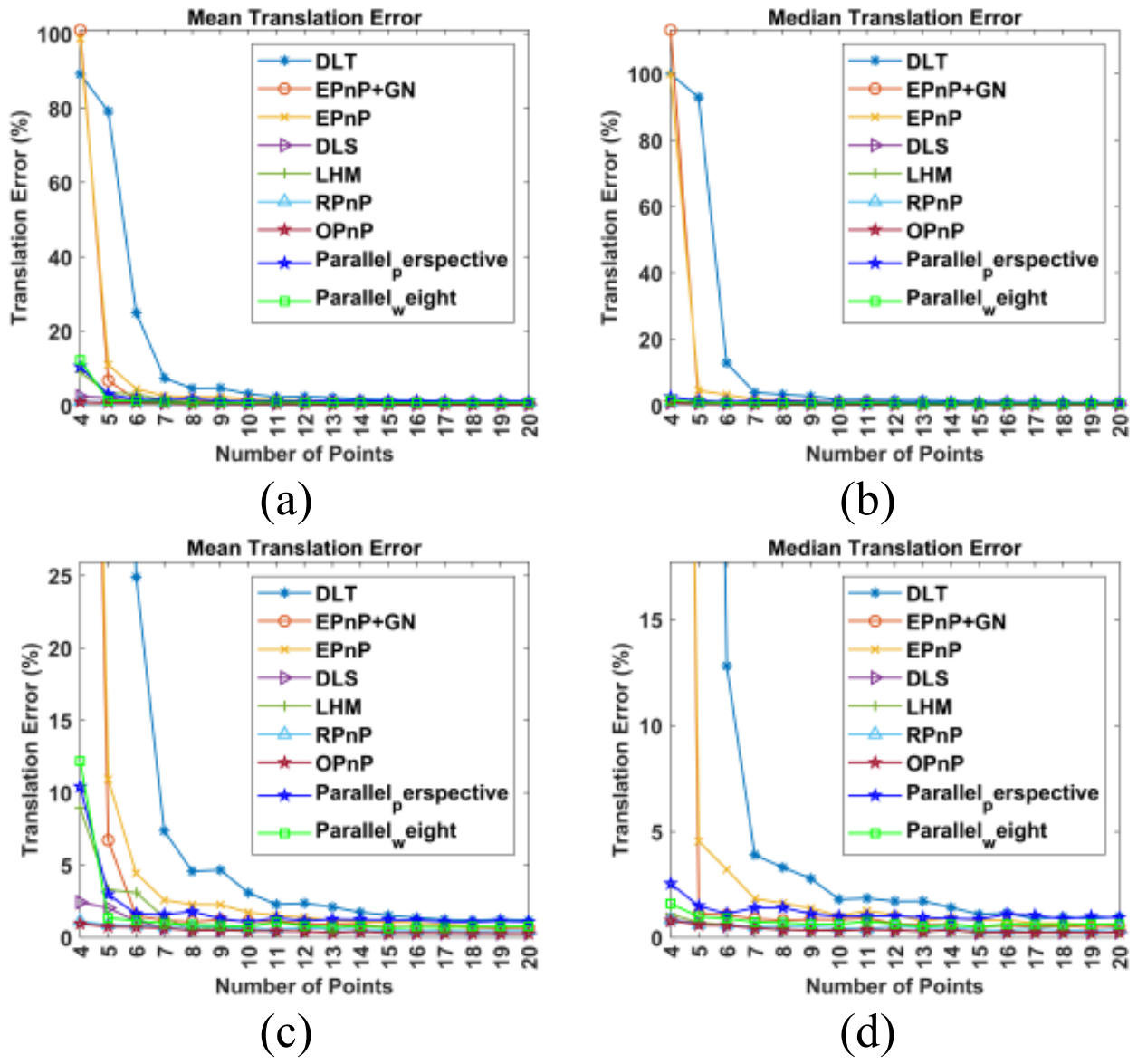

**Fig. 7b.** Effects of the number of feature points on the accuracies of position estimates ((a) and (b) are global plots, while (c) and (d) are local plots).

(1) As the number of feature points increases, the attitude estimation accuracies of both the nonlinear iterative algorithm and the analytic solving algorithm improve. Figs. 7a and b show that when the number of marker points is ≥ 12, both the iterative solution and the analytical solution tend to stabilize, the attitude estimation accuracy is within 0.7%, and the position estimation error is within 1.4%.

(2) The rotation and translation errors of the DLT and EPnP algorithms are poor overall, mainly because some nonlinear constraints are ignored and the fact that these algorithms are sensitive to noise, especially in cases with fewer feature points, which leads to relatively poor attitude estimation accuracy.

(3) LHM is a typical iterative algorithm with high algorithmic accuracy. The weighted parallel perspective estimation algorithm proposed in this paper has higher accuracy than LHM does, especially when few feature points are present (n<6), and the performance of the weighted parallel perspective estimation algorithm developed in this paper is only slightly worse than that of RPnP and OPnP.

(4) The weighted parallel perspective algorithm proposed in this paper uses the original parallel perspective algorithm to provide a relatively accurate initial position and then uses the weighted Gauss–Newton algorithm for optimization purposes to improve the accuracy of the attitude estimate. From Fig. 7, the weighted algorithm developed in this paper is slightly better than the parallel perspective attitude estimation algorithm without Gaussian optimization.

- Impact of noise on algorithmic accuracy

This experiment examines the noise sensitivity predicted by the proposed error transfer model, which characterizes how image-space measurement noise is amplified through perspective geometry under varying feature spacing and camera depth.

This experiment verifies the effect of noise on the tested algorithms. In the simulated data experiment, the number of feature points is fixed at 10 and 6, and the point data are obtained according to the simulated data generation method described in the previous section. The focal length of the camera is chosen to be 80 mm, and 500 cycles are chosen to add white noise with a variance of 0~10 pixels (with a step size of 0.5 pixels) to each reference point to verify the accuracy of the algorithms. The results of the experiments are shown in Figs. 8, 9a and b.

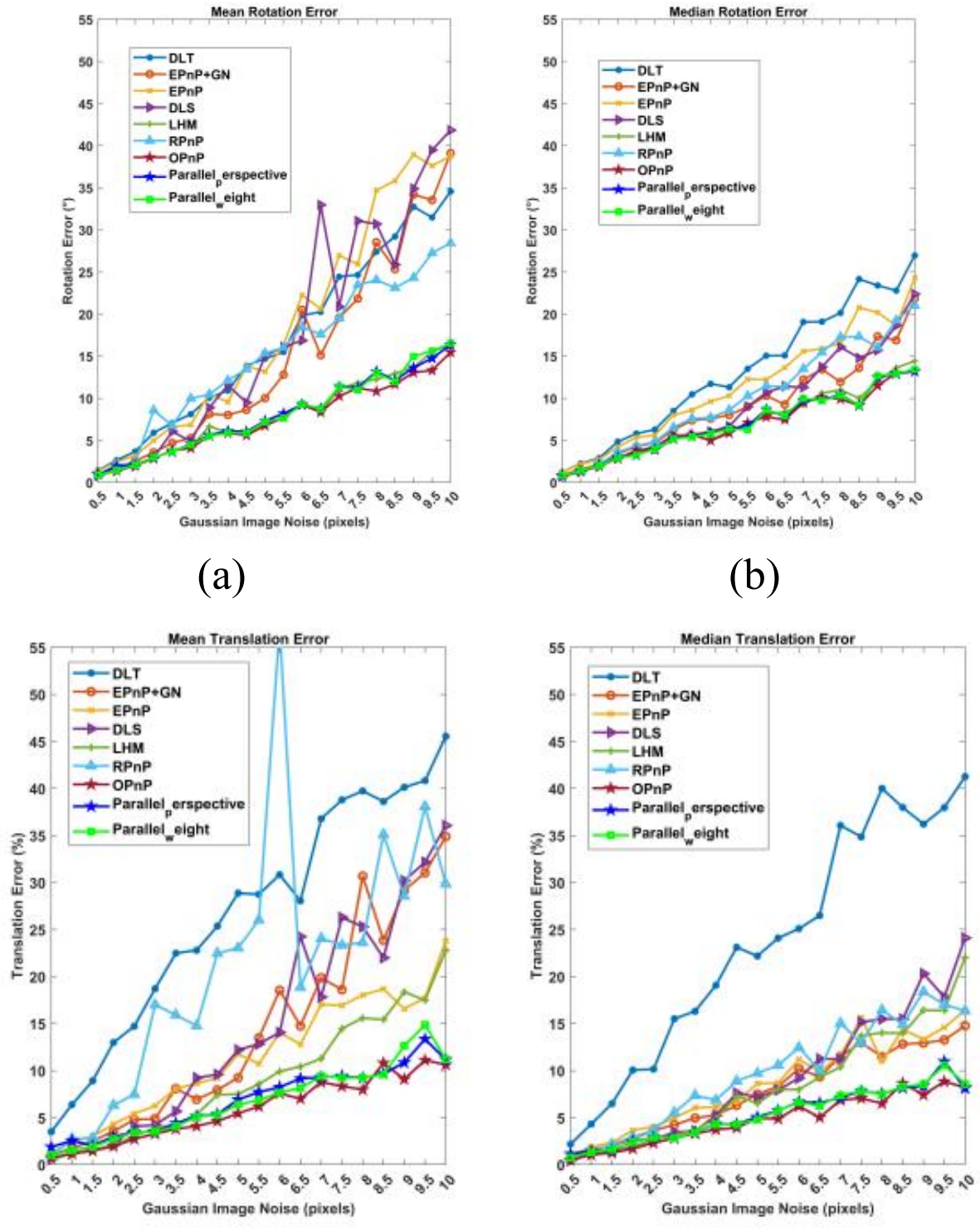

(a) (b)

(c) (d)

**Fig. 8.** Effects of utilizing 10 feature points and different noise magnitudes on the accuracies of the position estimation algorithms. ((a) and (b) are local plots, while (c) and (d) are global plots)

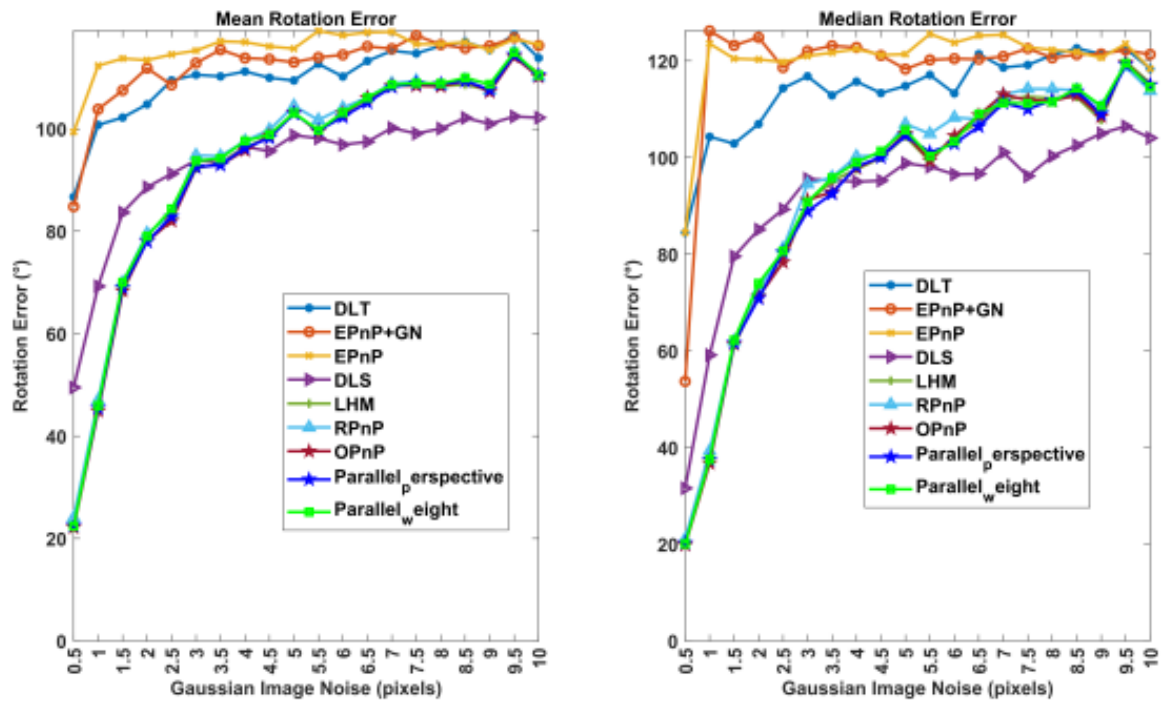

(a) (b)

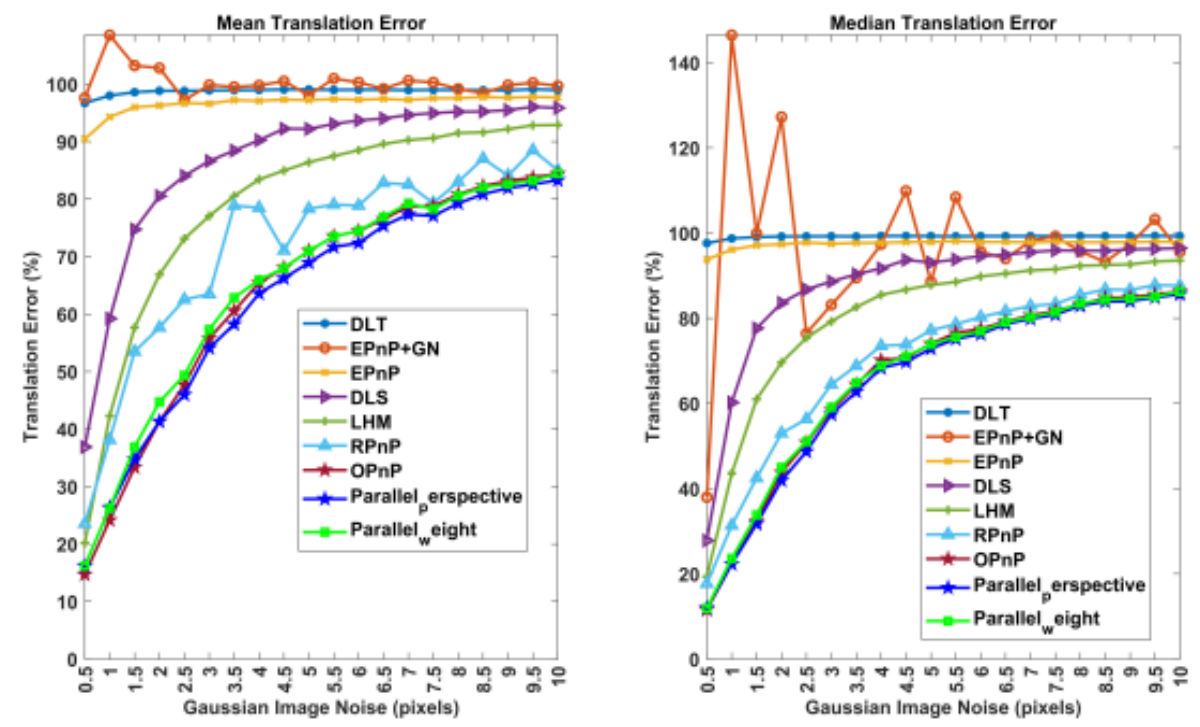

(c) (d)

**Fig. 9a.** Effects of utilizing 6 feature points and different noise magnitudes on the accuracies of the position estimation algorithms under an edge length of 25. ((a) and (b) are local plots, while (c) and (d) are global plots)

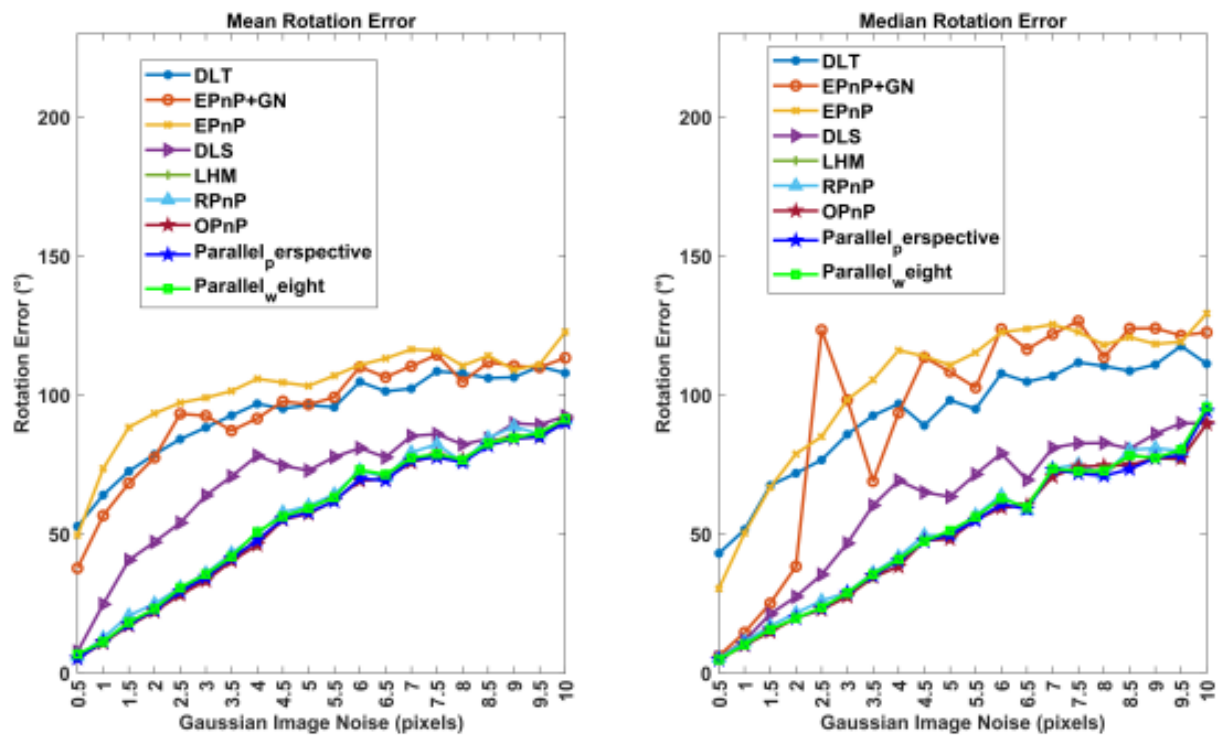

(a) (b)

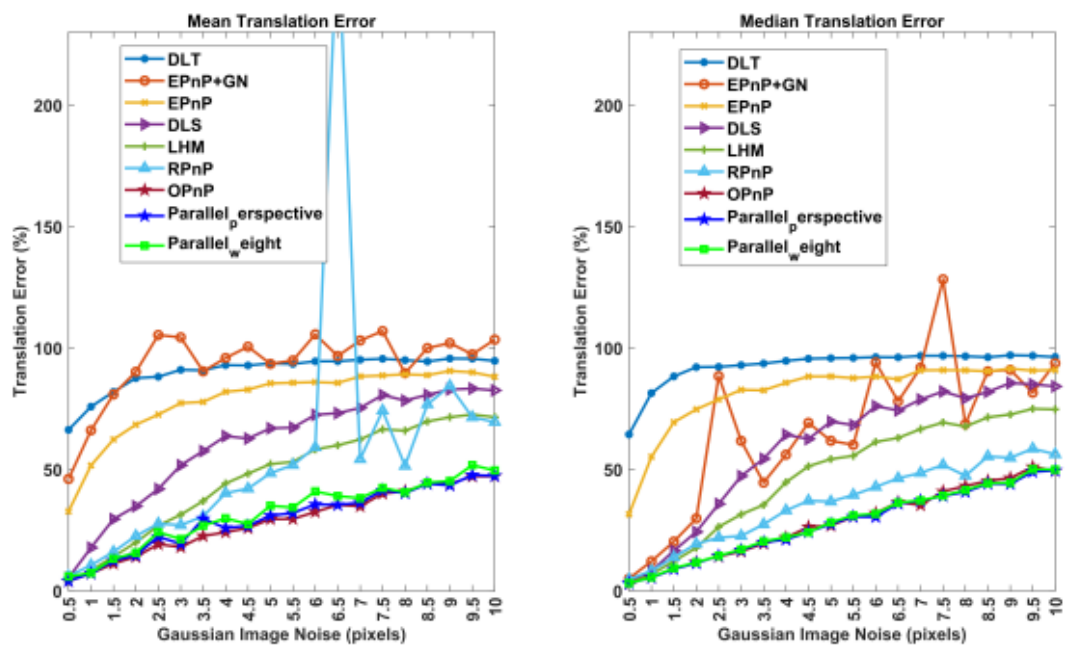

(c) (d)

**Fig. 9b.** Effects of utilizing 6 feature points and different noise magnitudes on the accuracies of the position estimation algorithms with an edge length of 100. ((a) and (b) are local plots, while (c) and (d) are global plots)

(1) The errors of both the iterative and analytic solution algorithms increase with increasing noise. The algorithms examined in this paper, such as OPnP, RPnP, and LHM, exhibit better resistance to noise, and the best accuracy can be stably obtained regardless of the distance between the marked points.

(2) The distance between the marker points affects the interference effect of noise; the larger the distance is, the greater the accuracy of the tested algorithm. The error curve of each algorithm tends to be a straight line, and the accuracy is improved as a whole. Fig. 8 shows that 10 marking points take the value range of 500 mm × 500 mm × 1000 mm, and the error curve of each algorithm is a straight line, with a maximum noise error of 15.7%, a panning error of 12%, a minimum noise error of 0.77%, and a panning error of 0.7%. Fig. 8 shows that the errors induced by the more accurate algorithms (i.e., the OPnP, RPnP, parallel perspective, parallel weight, and LHM algorithms) at a distance of 100 mm from the 6 feature points are straight lines, with a maximum noise attitude error of 84.6% degrees, a maximum panning error of 55.2%, a minimum noise error of 5.35% degrees and a minimum panning error of 3.7.6%. The feature points are curved for each algorithm at a distance of 25 mm, with a maximum of 112% degrees when panning by 85.6% and a minimum of 24% when panning by 17.7%. The size of the target object can effectively improve the robustness of each algorithm against noise.

(3) OPnP is the most accurate analytic solution algorithm for conducting bit position estimation with good noise immunity and is even better than the typical iterative algorithm (LHM) [revisiting PnP in a fast generalized solution case]. The weighted parallel perspective algorithm proposed in this paper is equivalent to the OPnP algorithm in terms of its robustness against noise interference.

- Operational efficiency of the algorithms

This experiment evaluates the computational implications of incorporating geometry-consistent error-aware weighting, verifying that the proposed robustness improvements do not incur prohibitive computational overhead.

This experiment verifies the running efficiency levels of the tested algorithms and compares them: In the simulated data experiment, the random point selection range is ±100 mm× ±100 mm× ±200 mm. Within the range of 4~104, using step 4, the number of reference points is changed, the experiment is repeated 500 times under different numbers of reference points, and then the average running time of each algorithm is determined. The experimental results are shown in Fig. 10, through which the following conclusions can be drawn.

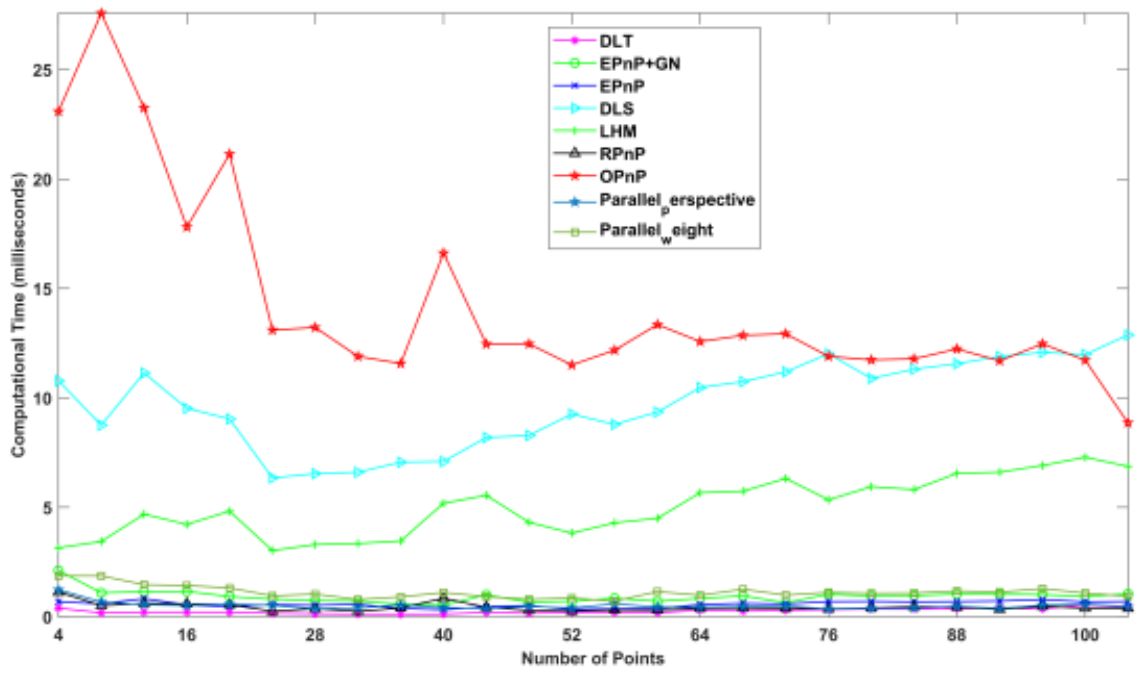

**Fig. 10.** Effect of the operational efficiency of the algorithms.

(1) The EPnP-based parsing solution has excellent computational efficiency, and a change in the number of feature points has little effect on the computational efficiency of this method. The LHM and DLS algorithms exhibit nearly linear running time increases with an increasing number of feature points.

(2) Among the algorithms developed for parsing solutions, the computational time of the OPnP algorithm is greater than those of the other algorithms and is approximately five times greater than those of the other parsing solutions. This is mainly due to the use of the Grobnacki solver, which increases the imposed computational cost; however, the computational cost is not related to the number of feature points.

(3) The running time of the weighted optimized parallel perspective projection algorithm proposed in this paper is close to that of other analytical solutions, such as RPnP, DLT, EPnP and EPnP+GN, and slightly greater than that of other analytical methods. The computational efficiency of the parallel perspective method without optimization is the same as that of EPnP, which indicates that the weighting process of the proposed algorithm affects the computation speed. The results show that the algorithm developed in this paper has a more optimized running efficiency level while maintaining high accuracy.

- Influence of the distance between the camera and feature point coordinate system on the algorithms

This experiment directly evaluates the depth-dependent error amplification behavior predicted by the proposed error propagation model, which identifies camera–target distance as a dominant factor governing near-field pose instability.

This experiment is aimed at verifying the effect of the distance between the camera and the feature point coordinate system on the tested algorithms. In the simulated data experiment, the number of feature points is fixed at 10 and 6, thus defining the ratio of the distance between the focal point of the camera and the origin of the feature point coordinate system to the distance between the feature points as the camera-feature point coordinate system depth-size ratio. Since the number of feature points is 10, the origin of the feature points and the distances between the feature points change. Therefore, the camera-feature point coordinate system depth-size ratio is defined as the ratio of the distance from the camera focus to the smallest edge of the feature point selection range to the size of the feature point selection range. The focal length of the camera is set to 500, white noise with a variance of 0.2 is added, 100 cycles are implemented in the area close to the focal point (0~500 mm from the focal point, with a step size of 100 mm) and away from the focal point (0~10000 mm from the focal point, with a step size of 500 mm), and the depth-size ratio of the camera-feature point coordinate system is changed to validate the accuracy of the position estimate produced by each algorithm. The results are shown in Figs. 11-14. The following conclusions can be drawn from the analysis.

(1) As the camera-feature point coordinate system depth-size ratio increases, the errors of the algorithms also increase and become linear. The DLT and EPnP algorithms are most affected by the camera-feature point coordinate system depth-size ratio.

(2) When 10 feature points are randomly selected, the selection range is large, the number of points is large, the depth–size ratio of the camera–feature point coordinate system is less than 20, and the attitude error has better linear characteristics. The performance of the proposed algorithm is consistent with that of the DLS, RPnP, and LHM algorithms, and it has good accuracy, with a minimum attitude accuracy of 0.5 degrees and a minimum displacement accuracy of 0.4%.

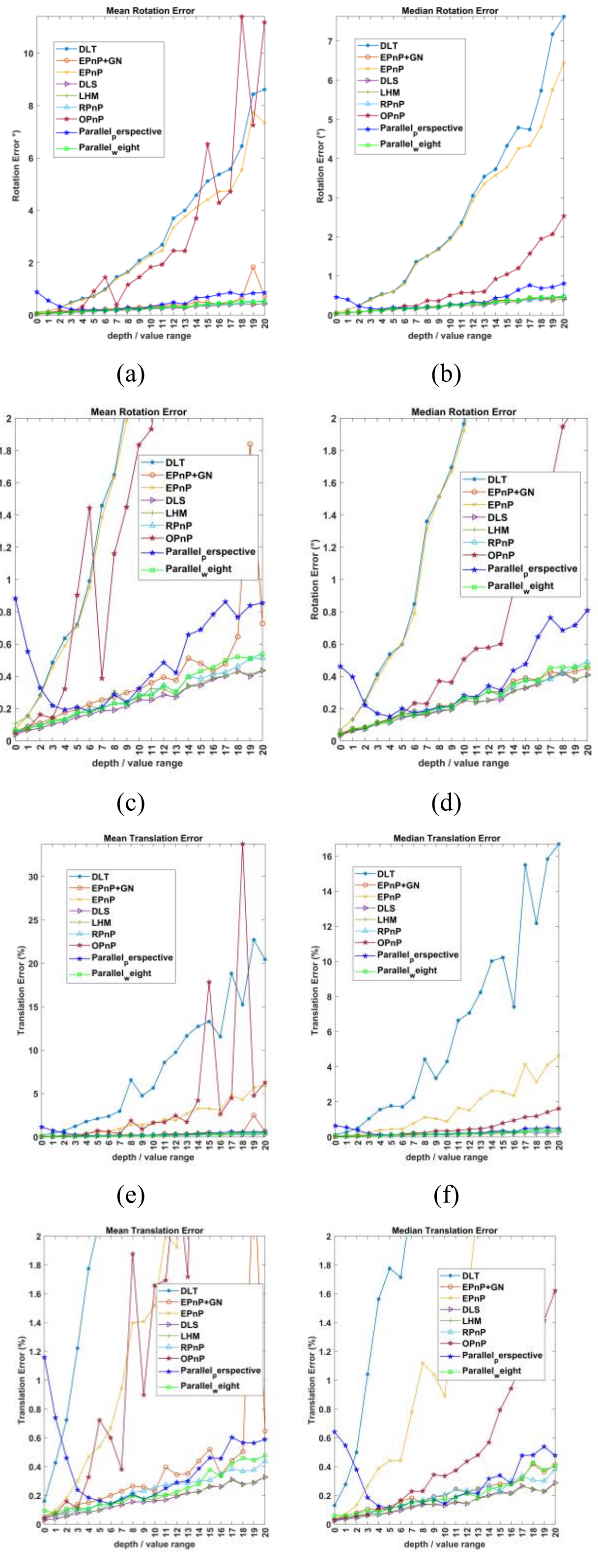

(a) (b)

(c) (d)

(e) (f)

(g) (h)

**Fig. 11.** Effects of the depth-size ratio of the feature point coordinate system of the camera on the accuracies achieved by the algorithms for 10 reference points.((a) and (b) are global plots, while (c) and (d) are local plots;.(e) and (f) are global plots, while (g) and (h) are local plots; )

(3) When the number of feature points is 6, the layout of the feature points is fixed, the distance between each pair of feature points is small, and the depth-feature point target size ratio is large, reaching 200. The algorithms examined in this paper and the parallel perspective LHM, RPnP, and OPnP algorithms have good immunity to changes in the depth-size ratio of the camera-feature point coordinate system, and the error curves exhibit flat and increasing trends. The error curves of DLT (under the 6-feature-point condition only), as well as the error curves of the EPnP, EPnP+GN and other algorithms, are characterized by two straight lines, with the inflection point at the position where the depth-feature point target size ratio is approximately 40. The large error variation observed in the first stage shows that in the camera proximity stage, the algorithmic accuracy is strongly affected by the variation exhibited by the depth-feature point coordinate system size ratio, and the influence of the variation in the camera-to-feature-point coordinate system–depth-to-feature-point coordinate system size ratio decreases after a certain value is reached.

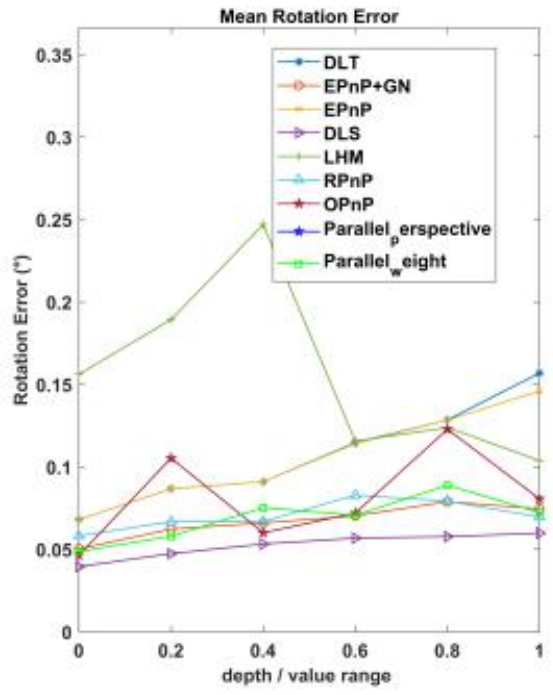

(a)

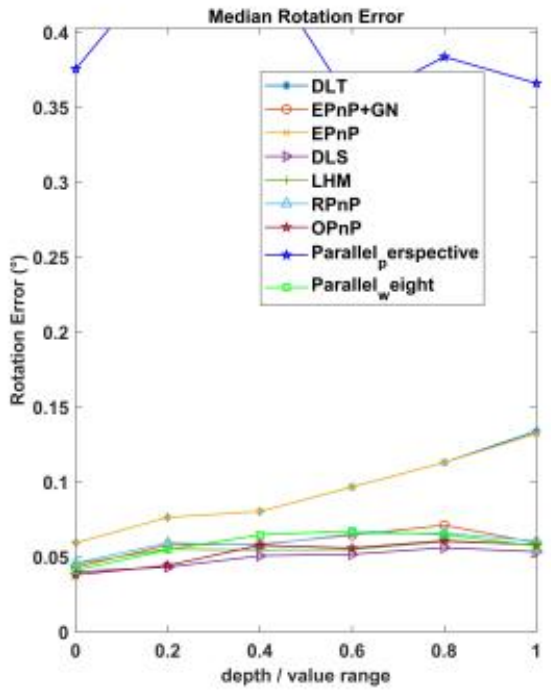

(b)

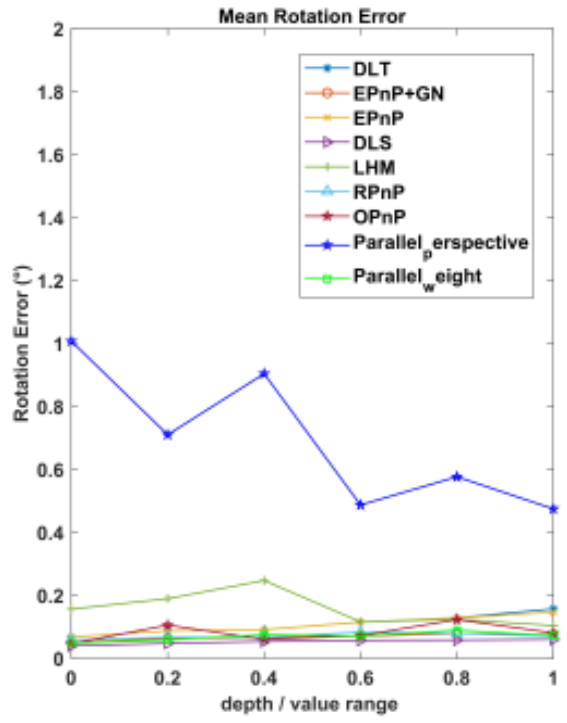

(c)

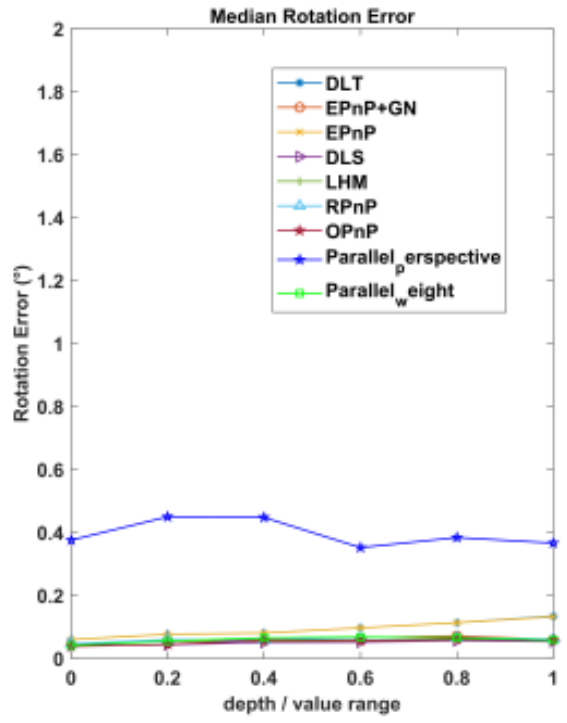

(d)

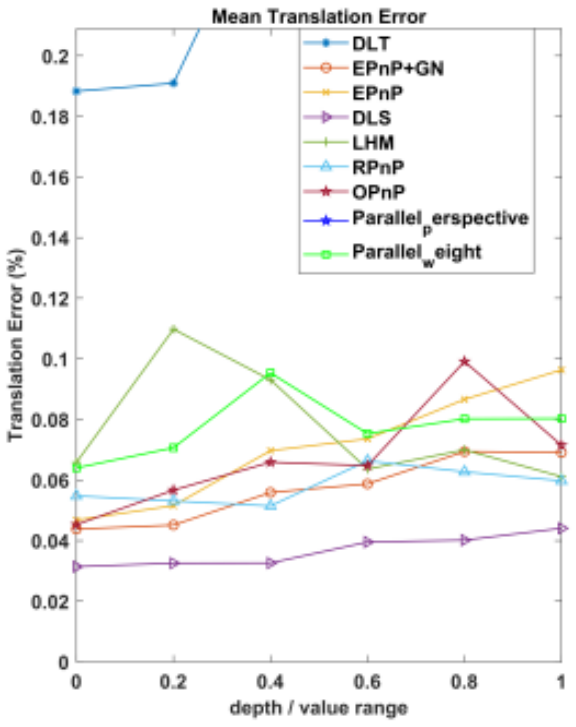

(e)

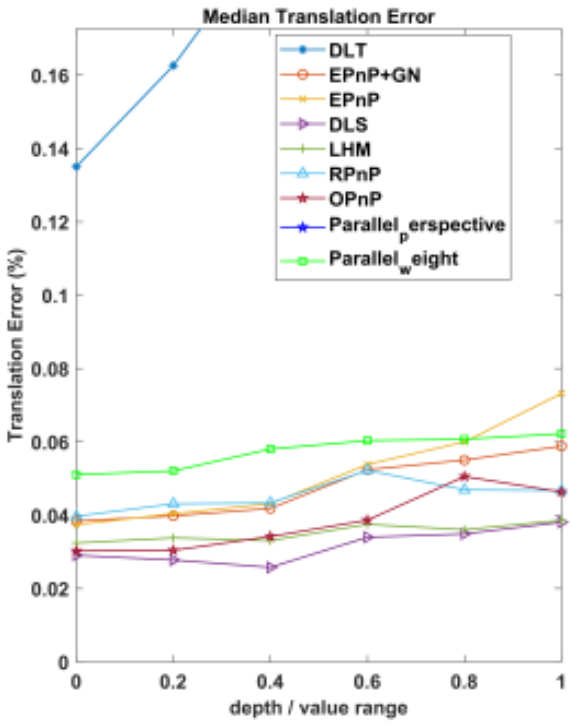

(f)

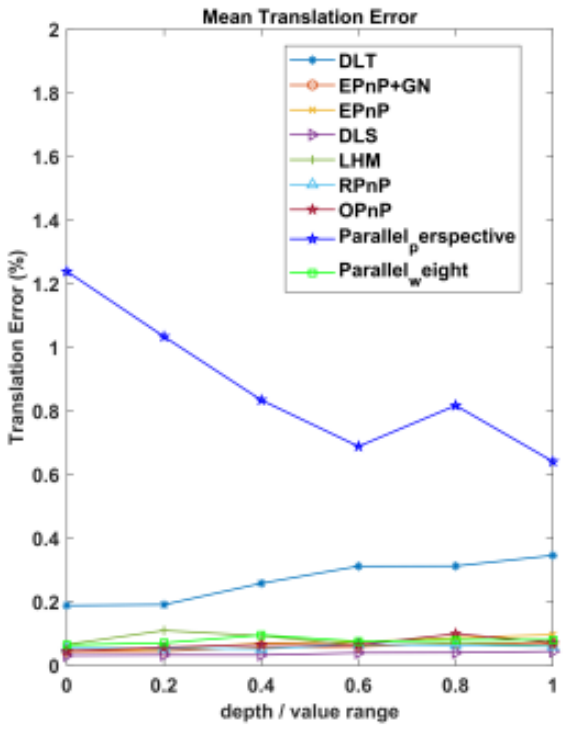

(g)

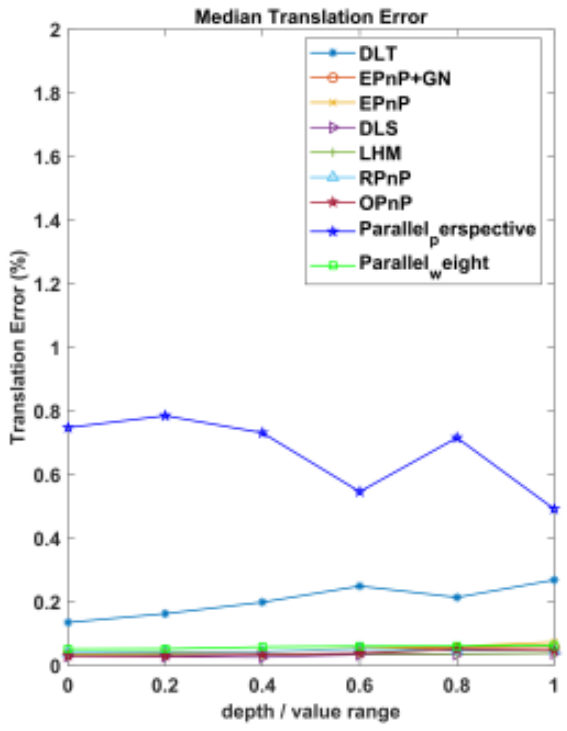

(h)

**Fig. 12.** Impacts of a camera depth/reference point ratios between 0 and 1 on the accuracies achieved by the algorithms for 10 reference points. ((a) and (b) are local plots, while (c) and (d) are global plots;.(e) and (f) are local plots, while (g) and (h) are global plots; )

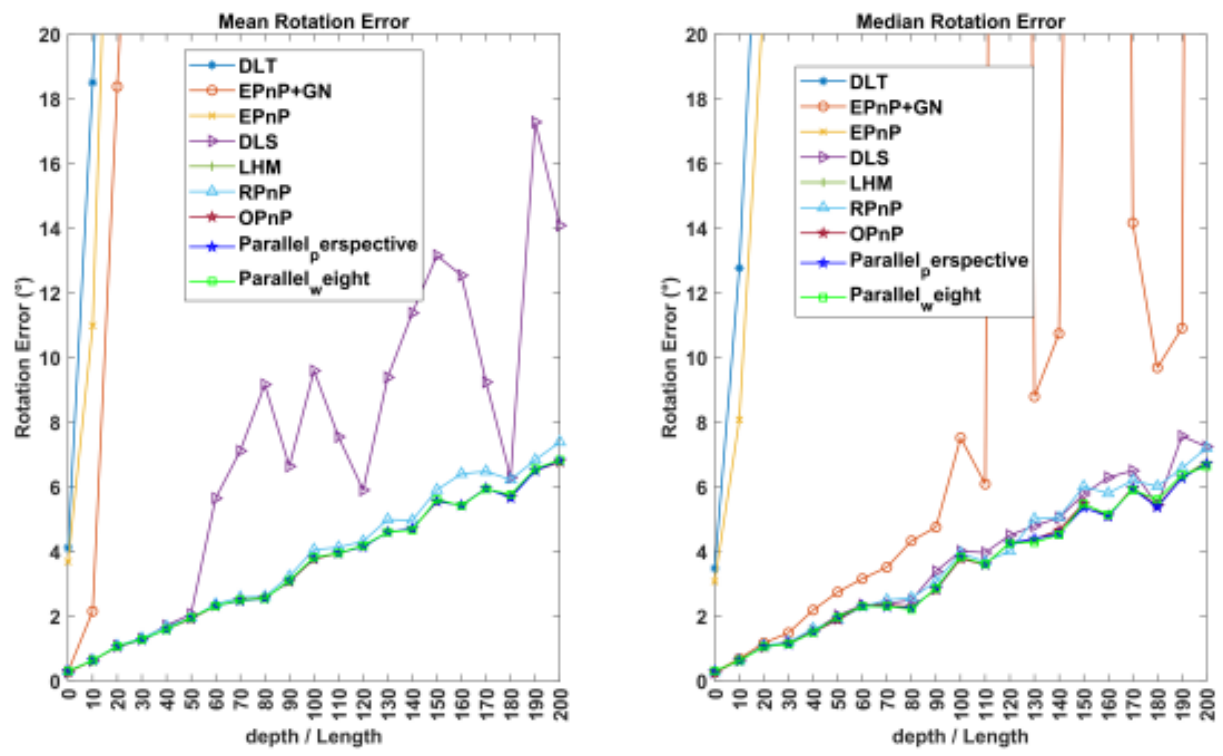

(a) (b)

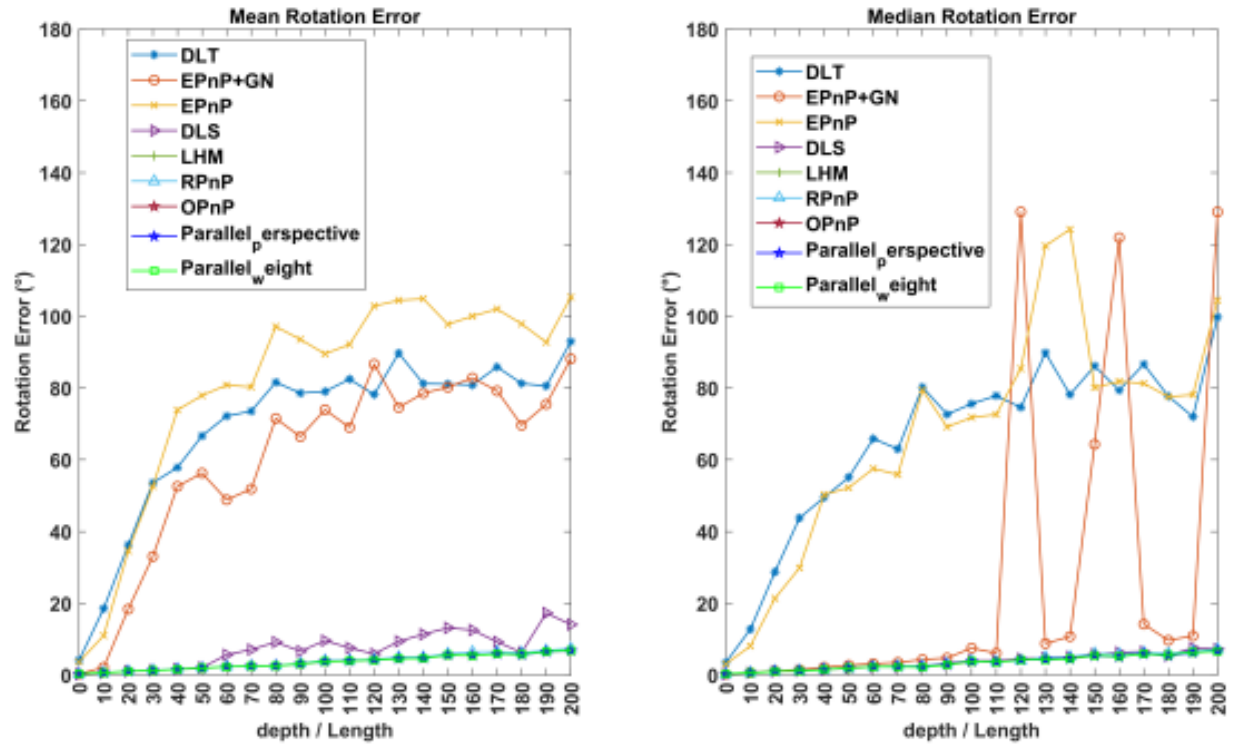

(c) (d)

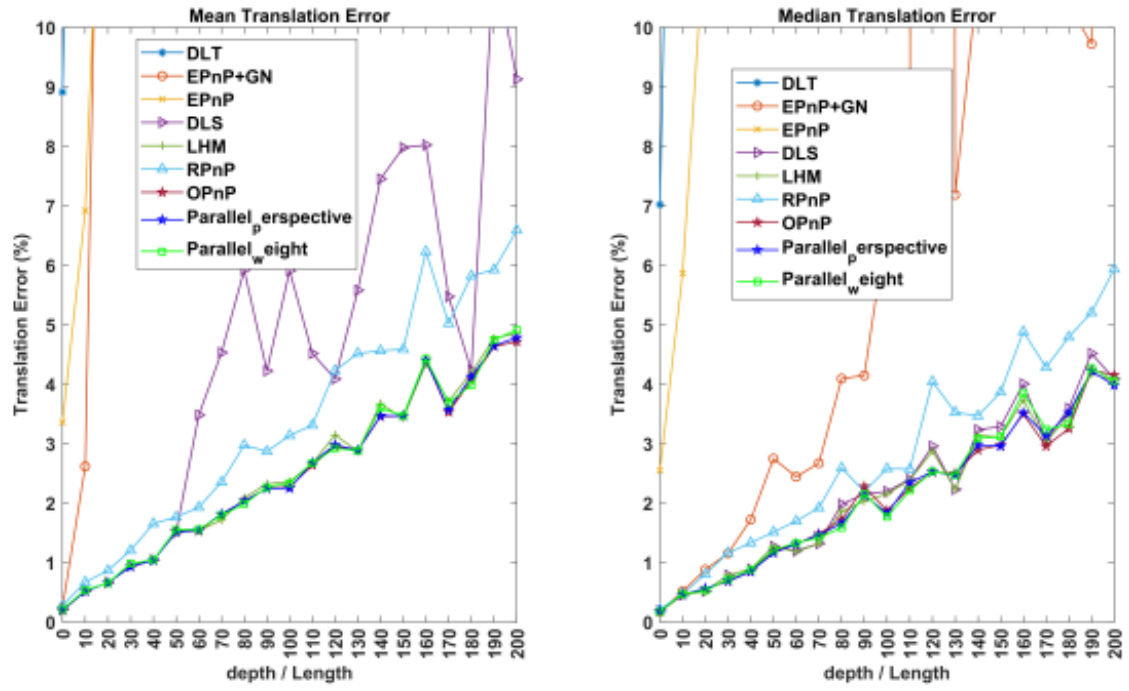

(e) (f)

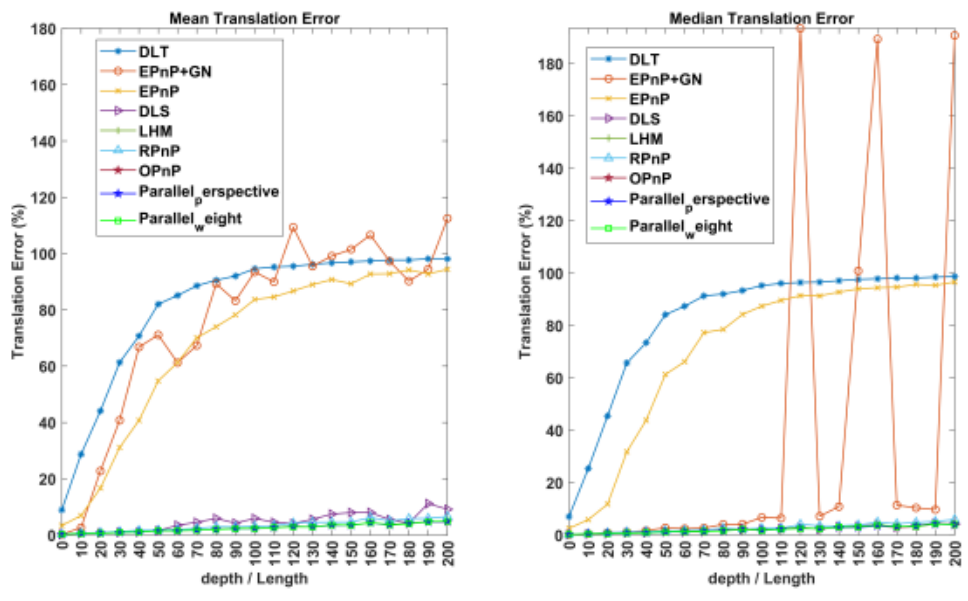

(g) (h)

**Fig. 13.** Effects of the depth–size ratio of the feature point coordinate system of the camera on the accuracies achieved by the algorithms for 6 reference points. ((a) and (b) are local plots, while (c) and (d) are global plots;.(e) and (f) are local plots, while (g) and (h) are global plots;)

(4) The initial number of feature points in the proposed algorithm is based on the parallel perspective projection model; because the parallel perspective projection model is an approximation algorithm involving perspective projection, especially when the feature points are close to the origin of the coordinate system and far from the optical axis, the initial value has a large error. As shown in Fig. 12, the error curve of the parallel perspective projection algorithm has a large error in the range with depth-feature point target size ratios that are less than 4. The weighted optimization strategy developed in this paper effectively eliminates this error and improves the accuracy of the algorithm. This shows that the algorithm proposed in this paper can cope well with the interference caused by changes in the depth-to-feature point coordinate system size ratio and is suitable for a wider range of application scenarios.

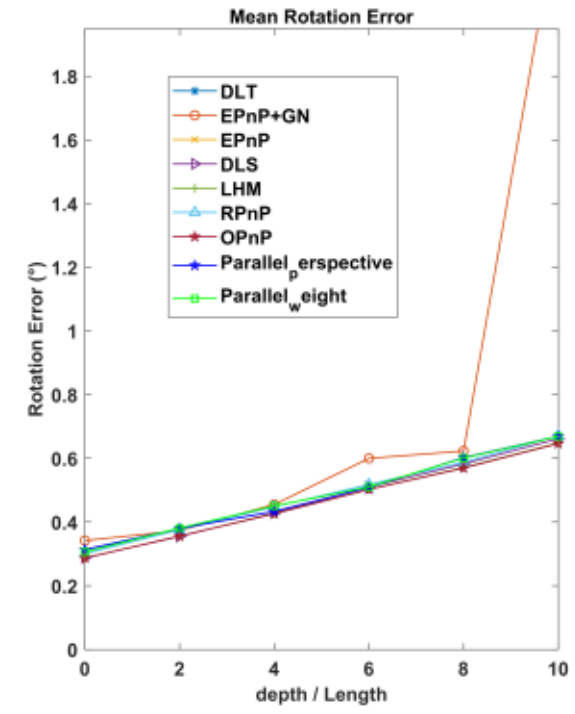

(a)

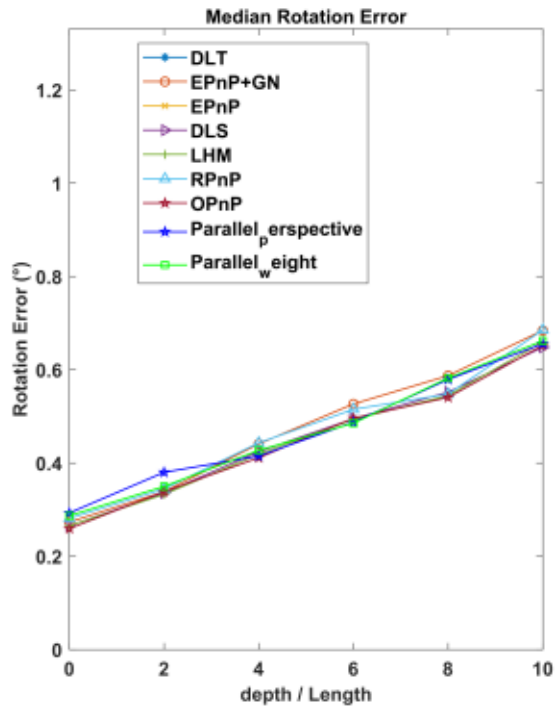

(b)

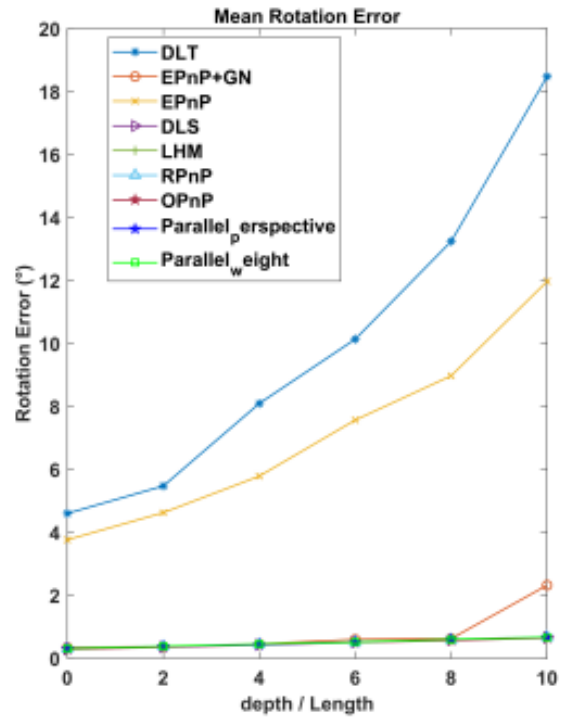

(c)

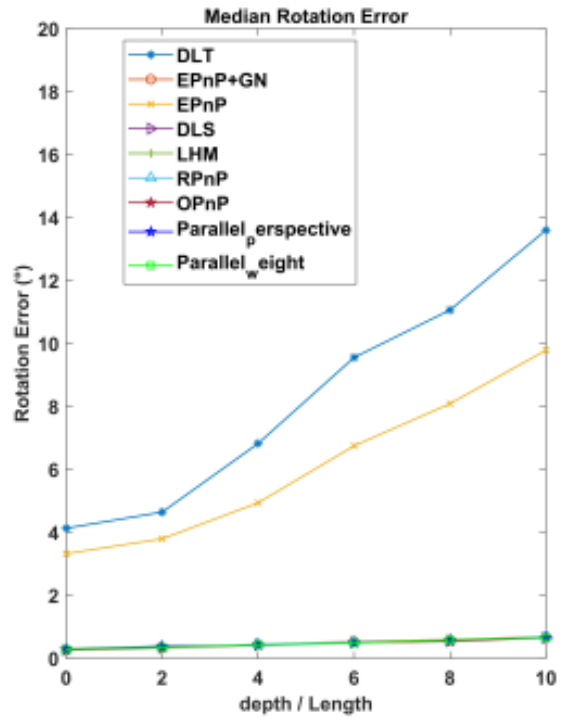

(d)

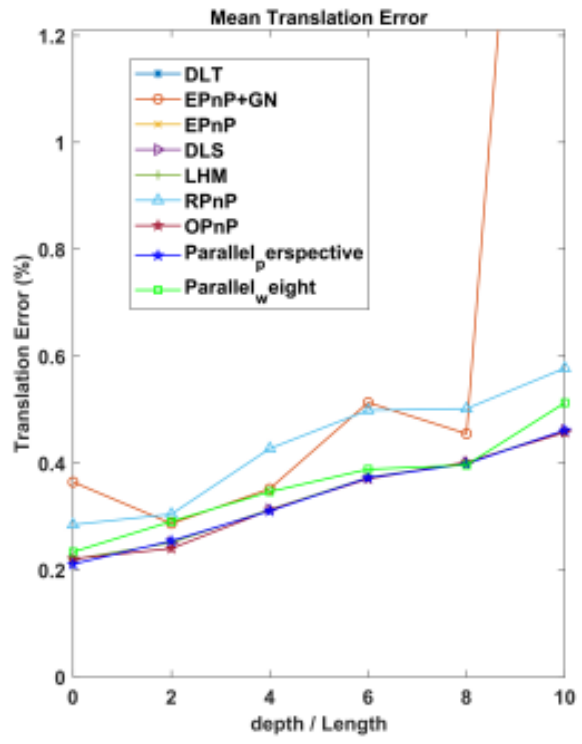

(e)

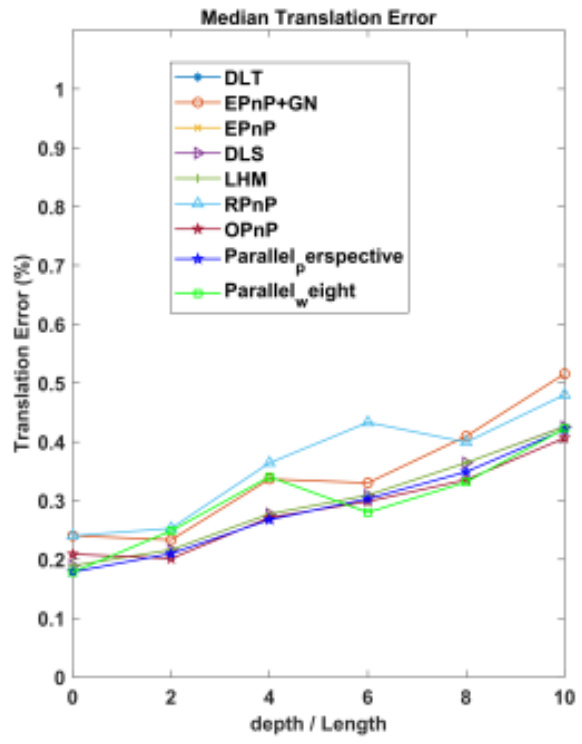

(f)

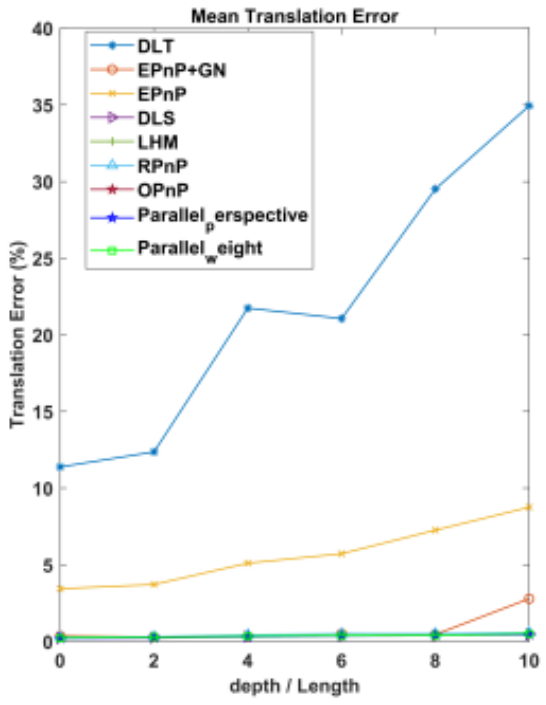

(g)

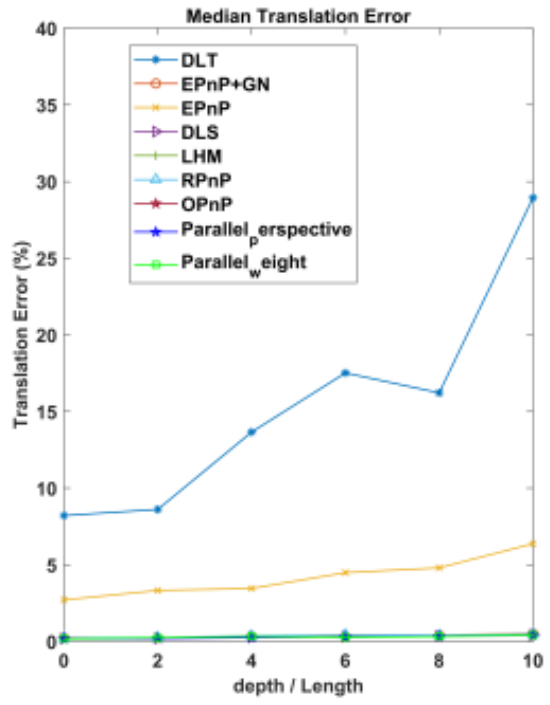

(h)

**Fig. 14.** Impacts of camera depth/reference point ratio values between 0 and 1 on the accuracies achieved by the algorithms for 6 reference points.((a) and (b) are local plots, while (c) and (d) are global plots;.(e) and (f) are local plots, while (g) and (h) are global plots; )

- Effect of the distance from the origin of the feature point coordinate system to the optical axis of the camera on the algorithms

This experiment validates the influence of feature eccentricity relative to the optical axis, as predicted by the error transfer analysis, which shows that off-axis configurations induce structured pose estimation bias in near-field scenarios.

This experiment verifies the effect of the distance from the origin of the feature point coordinate system to the optical axis of the camera on the algorithms. In the simulated data experiment, the number of fixed feature points is 6 (with a stereoscopic distribution), the focal length of the camera is 80, white noise with a variance of 0.2 is added, the number of cycles is chosen to be 100, the X-axis and Z-axis coordinates of the origin of the fixed feature point coordinate system are fixed at 0, and the Z-axis coordinates are 200 mm away from the focal point and 100 mm from the focal point. The origin of the Y coordinate parameter is changed to 1~500 mm, the step size is 25 mm, and different distances from the origin of the feature point coordinate system to the optical axis of the camera are simulated to verify the solution accuracy of each algorithm. The experimental results are shown in Figs. 15 and 16. The horizontal axes of Supplementary Fig. S1 and Supplementary Fig. S2 denote the ratio of the distance from the origin of the feature point coordinate system to the optical axis of the camera to the distance between the feature points, and the following conclusions can be drawn from the analysis.

(1) The distance from the origin of the feature point coordinate system to the optical axis of the camera affects the accuracy of the initial feature point positions of different algorithms, thus affecting the accuracy of the pose estimates yielded by the different algorithms. The DLT and EPnP algorithms are most affected by the influence of the distance from the optical axis, and their pose estimation errors are large.

(2) The algorithms proposed in this paper and the OPnP, RPnP, LHM, DLS, and EPnP+GN algorithms have high levels of resistance to the interference of the eccentricity distance change, the attitude errors are all between 0.2 and 0.°, and the displacement errors are all between 0.5% and 0.8%.

(3) The initial value of the feature point of the proposed algorithm is calculated via the parallel perspective model, and the deviation distance from the optical center to the origin of the feature point coordinate system and the proximity distance to the focal point affect the accuracy of the initial value. As seen from the proximity error map, the weighted optimization strategy proposed in this paper can improve the accuracy of parallel perspective estimation. When the ratio of the distance from the optical axis of the camera to the distance between the feature points is 9.52, the camera pose error is reduced from 2.2° to 0.62°, which is a 3.5-fold improvement, and the movement error is reduced from 0.67% to 0.47%, representing a 1.42-fold improvement.

- The effect of the angle between the optical axis of the camera and the Z-axis of the reference point coordinate system on the algorithms

This experiment examines the angular dependency of pose estimation error predicted by the proposed error propagation model, which relates attitude sensitivity to the relative orientation between the feature coordinate system and the camera optical axis.

As shown in in Supplementary Fig. S3, this experiment verifies the effect of the angle between the optical axis of the camera and the Z-axis of the reference point coordinate system on the algorithms. In the simulated data experiment, to verify the effect of the angle between the marker point coordinate system and the optical axis on the positional attitude estimates, the number of fixed feature points is 6, the origin is fixed at (0, 0, 500), the angles around the X-axis and Z-axis are 0, and the angle of the reference point coordinate system rotating around the Y-axis is varied (-90~90, step 1). The focal length of the camera is 500 mm, white noise with a variance of 0.2 is added, and 100 cycles are chosen to verify the solution accuracy of the algorithms. The following conclusions can be drawn from the analysis.

(1) The attitude of the feature point coordinate system has different effects on the attitude estimation accuracies of different algorithms. The DLT, EPnP, and EPnP+GN algorithms have large position attitude estimation errors, and each position attitude error curve has a double U shape, with a peak in the region near 0 degrees; in contrast, the attitude estimation accuracies of the other algorithms are not affected by the attitude of the feature point coordinate system. The accuracies of the algorithm proposed in this paper and the OPnP algorithm are the best.

(2) According to (29), the attitude error estimated by the algorithm proposed in this paper is related to the angle between the Z-axis and the optical axis of the coordinate system of the characteristic point $\beta$ with $1/(\sin\beta(\cos\beta)^{0.5}$ ). The inflection point of its error curve is located at the inflection point of the function f=$1/\sin\beta(\cos\beta)^{0.5}$

$$\left(\frac{1}{\sin\beta(\cos\beta)^{0.5}}\right)' = 0$$

$$\beta = 54.7^{\circ}$$

The attitude error of the proposed algorithm is between 0.5° and 0.7°, and the movement error is between 0.3% and 0.5% (straight). This shows that the stereo feature point layout utilized in this paper has high application value.

*2) Real Image Experimental Results and Analysis*

This set of real-image experiments evaluates the practical validity of the proposed geometric error propagation framework under heterogeneous sensing conditions. In particular, the experiments are designed to verify whether the depth-dependent error amplification, feature eccentricity effects, and noise-induced instability predicted by the error transfer analysis persist in real-world near-field scenarios.

Under different lighting environments, the robot moves while carrying the feature point coordinate system and then rotates around the X-axis, starting from 0 degrees and rotating by 2 angles each time until it reaches ±90°; a picture is taken after each rotational movement to measure the pose of the marker point coordinate system and calculate the error according to (33). Considering that the bit position measurement of each motion position is based on one picture, the noise interference level is high, which affects the evaluation results of different algorithms. Therefore, in this paper, the average value of the bit pose errors at different rotation angles from the moving position of the feature point coordinate system is used to evaluate the accuracy and robustness of different algorithms. The average value of the errors of the angles at different positions is used to evaluate the accuracy and robustness of different algorithms.

The data in the experimental data table are the average camera pose estimation errors produced for the feature point coordinate system at different distances from the camera in the Z-axis direction of the camera. Each row contains the average camera position estimation errors produced by different position measurement algorithms for 90 different poses at the same location. Each column contains the average camera pose estimation errors produced by the same pose measurement algorithm for 90 different poses at different distances from the optical axis of the camera at the same camera depth.

Supplementary Material Tables I-V present the experimental data produced under the atmospheric spotlight environment, Supplementary Material Tables VI-X present the experimental data obtained under the surgical shadowless lamp environment, and Supplementary Material Tables XI-XV present the experimental data yielded under the underwater low-light environment. The following conclusions can be drawn from the analysis.

(1) The experimental data show that the distance between the camera and the marker point has a significant effect on the accuracy of each position estimation algorithm. As the depth increases (from 200 mm to 1000 mm), both the rotational and translational errors (T) increase significantly with the depth, both in the atmospheric bright light environment and in the surgical shadowless light environment. At 200 mm and 400 mm, most of the algorithms are able to maintain low errors with error values less than 2. However, when the depth increases to 1000 mm, the error values increases significantly, especially for the DLT and EPnP algorithms, which have the highest errors at up to 49.63° large error increases, indicating low stability. The reason for this phenomenon may be that the greater the depth is, the smaller the projection of the target feature points in the image, which makes it difficult for the camera to accurately estimate the attitude of the target, especially when the distance is large and the angle changes are large; thus, the error accumulates rapidly. In contrast, the parallel and parallel weighted algorithms are more adaptable to depth changes, and even at longer distances (e.g., 1000 mm), their error variations are smaller, resulting in greater stability and better resistance to the adverse effects of depth.

(2) Different lighting conditions significantly affect the accuracy of the position estimation algorithms. At 800 mm and 1000 mm, in the atmospheric bright light environment, because the depth is too large, it is difficult to extract the feature points from the picture, and the accuracy of the algorithms decreases. However, in the surgical shadowless lamp environment, owing to the better uniformity of the light, the feature points on the picture are clearly visible, the fluctuation exhibited by the error with the change in depth is relatively small, and the accuracy of the algorithms are more stable; this is especially true for the parallel and parallel weighted algorithms, whose rotational and translational errors do not change drastically, indicating strong robustness.

(3) In the underwater low-light environment, marker images are collected, the marker position information of these images is manually extracted, the position extraction error does not interfere much with the estimation accuracy of the proposed algorithm, and the accuracy of this algorithm is consistent with the accuracy of well-known algorithms, such as OPnP, LHM, RPnP, and DLS.

(4) Compared with those in a general environment, the lighting conditions and image quality in an underwater environment are more disturbed. Underwater images are generally blurred and noisy, resulting in poor algorithmic accuracy, especially with increasing distance, and the induced error increases significantly. Parallel and parallel weighted algorithms are more stable in underwater environments, and their errors are relatively small at most distances. In particular, the translation errors (TEs) are more accurate at closer distances, and the error increases less as the distance increases. The traditional DLT and EPnP algorithms perform poorly in underwater environments, especially at longer distances (e.g., 1000 mm), where the errors increase dramatically, especially the rotational error (RE) and TE) values; thus, these methods exhibit greater instability.

(5) The errors of the DLT and EPnP algorithms are strongly affected by the distance from the optical axis, and in the underwater environment, the RE errors are 5.75° (DLT) and 5.14° (EPnP) at a distance of 400 mm. The errors increase further when the optical axis is offset by 50 mm and 100 mm, which indicates that these two algorithms are very sensitive to changes in the offset optical axis. The algorithms examined in this paper are more stable and more resistant to changes in the error of the off-axis, with the rotational error varying within 0.7° and the translational error varying within 1% with the offset of the optical axis.

A comparison among the rows of data in the tables shows that the algorithm developed in this paper is one of the best and most stable algorithms in terms of estimation accuracy, regardless of the lighting environment, which is consistent with the conclusions of the simulated data experiments concerning noise interference. A comparison among the columns of data in the tables reveals that the further away the optical axis is, the more significant the change in the position estimation error, which is related to our error value being the average of the position measurement errors induced for 90 different poses. In the same lighting environment, a comparison among different tables reveals that the greater the distance from the camera is, the greater the algorithmic estimation error, which is consistent with the simulated data experiments concerning the interference of the distance variation between the camera and the feature point coordinate system.

# 5. Discussion

To address the inherent instability of camera pose estimation in near-field operations, this study examined the problem from a **geometric error propagation perspective**, rather than treating it solely as an algorithmic optimization task. By explicitly modeling how image-space measurement errors propagate through perspective geometry under varying camera–target distances and feature configurations, the proposed framework provides a structural explanation for the performance degradation commonly observed in analytical PnP methods in proximity scenarios. Within this framework, the parallel perspective approximation serves as a geometry-consistent initialization that mitigates depth-dependent bias, while the error-aware weighting strategy in the Gauss–Newton optimization explicitly accounts for heterogeneous uncertainty induced by perspective effects. As a result, the observed performance gains are not merely attributable to improved numerical optimization, but arise from aligning the pose estimation process with the underlying geometric structure of near-field perspective projection.

Extensive experimental evaluations on both simulated data and real-world images validate the theoretical insights derived from the proposed error transfer model. Across a wide range of camera–target distances, feature layouts, and illumination conditions—including atmospheric radiation, surgical lighting, and underwater low-light environments—the observed trends in rotational and translational errors closely follow the behavior predicted by the error propagation analysis. In particular, the consistent improvement over classical analytical methods such as DLT, EPnP, EPnP+GN, RPnP, and OPnP reflects the effectiveness of explicitly incorporating geometry-driven error characteristics into the estimation process. These results indicate that robustness in near-field pose estimation is fundamentally governed by error transfer mechanisms, and that explicitly modeling this structure leads to stable and computationally efficient performance across diverse operating conditions.

Importantly, unlike existing uncertainty-aware or globally optimal PnP solvers that primarily focus on improving the algebraic solution of the pose estimation problem, the proposed approach emphasizes **geometric interpretability**. By revealing how depth, feature distribution, and viewing configuration jointly influence error amplification in near-field scenarios, the proposed framework offers a principled explanation for why certain PnP formulations fail under proximity conditions and how such failures can be mitigated through geometry-consistent modeling.

### *5.1. Impact of the Feature Point Layout on Pose Estimation*

The number of feature points and their spatial layout play a critical role in determining the stability of PnP-based pose estimation. While six noncoplanar feature points with 3D–2D correspondences are sufficient to uniquely determine camera pose, increasing the number of feature points beyond this minimum can significantly improve estimation robustness. The experimental results demonstrate that both the proposed method and the comparison algorithms achieve stable and high pose estimation accuracy when the number of noncoplanar feature points exceeds thirteen.

From the perspective of error propagation analysis, this behavior can be attributed to the redundancy introduced by additional feature points, which reduces the influence of individual measurement errors on the overall pose estimate. The observed stabilization effect directly supports the theoretical prediction that pose uncertainty decreases as geometric constraints become more overdetermined, particularly in near-field regimes where perspective sensitivity is pronounced.

### *5.2. Weighting Strategies for Reducing Errors*

Most conventional PnP algorithms implicitly assume homogeneous measurement noise across feature points, an assumption that rarely holds in real-world near-field environments. In contrast, the proposed method incorporates a differential weighting strategy derived from the error transfer model, which accounts for factors such as camera–target depth, the distance of the feature point centroid from the optical axis, and inter-feature spacing.

By assigning lower weights to feature points with larger predicted uncertainty and emphasizing geometrically stable measurements, the Gauss–Newton optimization becomes more resilient to heterogeneous noise. Experimental results under simulated noise and varying depth-to-target size ratios confirm that the proposed approach achieves robustness comparable to high-performance methods such as OPnP and LHM. Notably, in close-proximity scenarios where the depth-to-target size ratio is less than four, the proposed method consistently outperforms the parallel perspective projection method without error-aware weighting, highlighting the importance of incorporating error propagation characteristics into the optimization process.

### *5.3. Feature Point Layout Optimization*

Leveraging the proposed error propagation framework, this study further analyzed the relationship between feature point distribution and pose estimation accuracy. Although a biprism feature layout was adopted for experimental validation, the theoretical analysis does not depend on a specific marker geometry. Rather, the derived error transfer relationships apply generally to noncoplanar feature configurations commonly encountered in near-field perception tasks.

The experimental results demonstrate that feature layouts designed in accordance with the error propagation analysis exhibit enhanced robustness to extraction errors, particularly under challenging sensing conditions such as space radiation, strong operating room lighting, and underwater low-illumination environments. These findings suggest that the proposed framework can serve as a practical guideline for marker and feature layout design in near-field vision systems beyond the specific configuration used in this study.

### *5.4. Real-World Performance*

Under diverse real-world lighting environments—including high-intensity atmospheric illumination, shadowless surgical lighting, and underwater low-light conditions—the proposed method maintains stable pose estimation performance. The accuracy of the estimated poses remains consistent with that of leading analytical and iterative algorithms such as OPnP, LHM, RPnP, and DLS, while exhibiting reduced sensitivity to depth variation and feature extraction degradation.

These observations further corroborate the theoretical claim that near-field pose estimation robustness is primarily governed by geometry-induced error propagation rather than by algorithmic complexity alone. By explicitly accounting for these geometric effects, the proposed approach achieves reliable performance across a wide range of operational conditions without sacrificing computational efficiency.

### *5.5. Future Research Directions*

While the present study employs manual feature point extraction to isolate and analyze geometric error propagation effects, practical deployment in real-time systems requires automated, fast, and robust feature extraction pipelines. Future work will focus on integrating the proposed error-aware pose estimation framework with advanced feature detection and tracking methods, as well as extending the analysis to dynamic targets and time-varying noise conditions. In addition, further exploration of geometry-driven error modeling in multi-camera and multi-sensor configurations represents a promising direction for enhancing robustness in complex near-field perception systems.

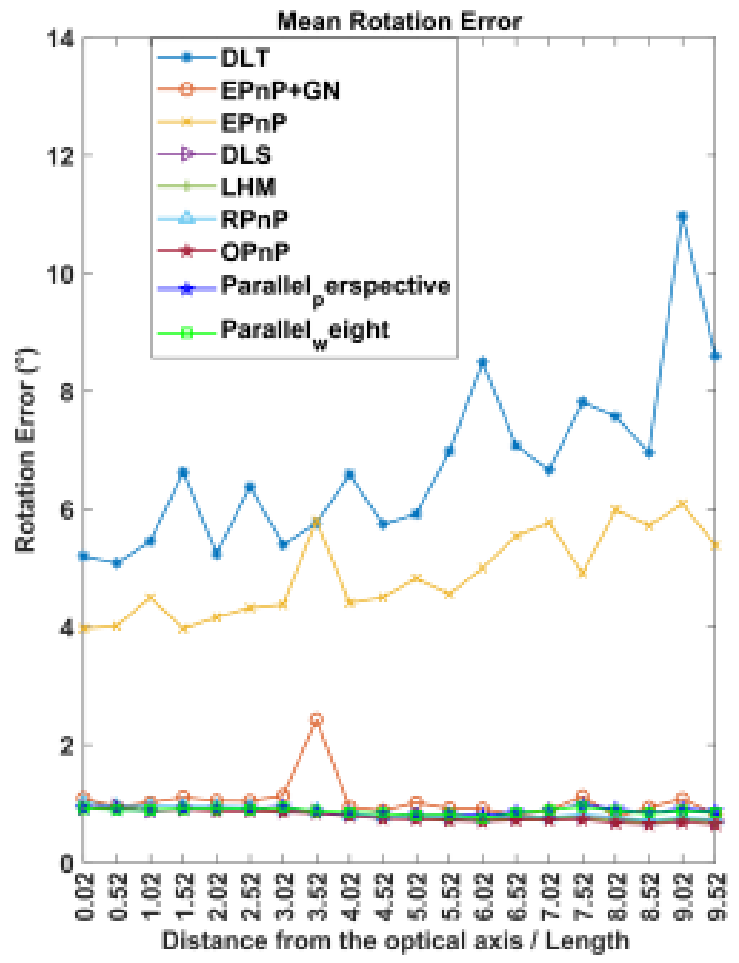

(a)

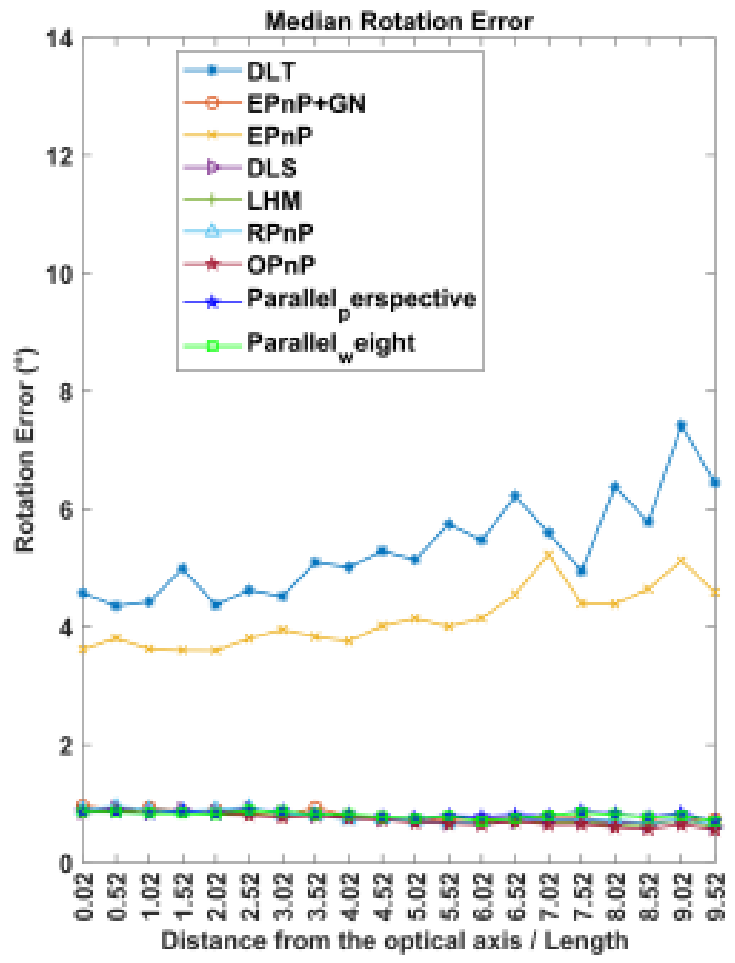

(b)

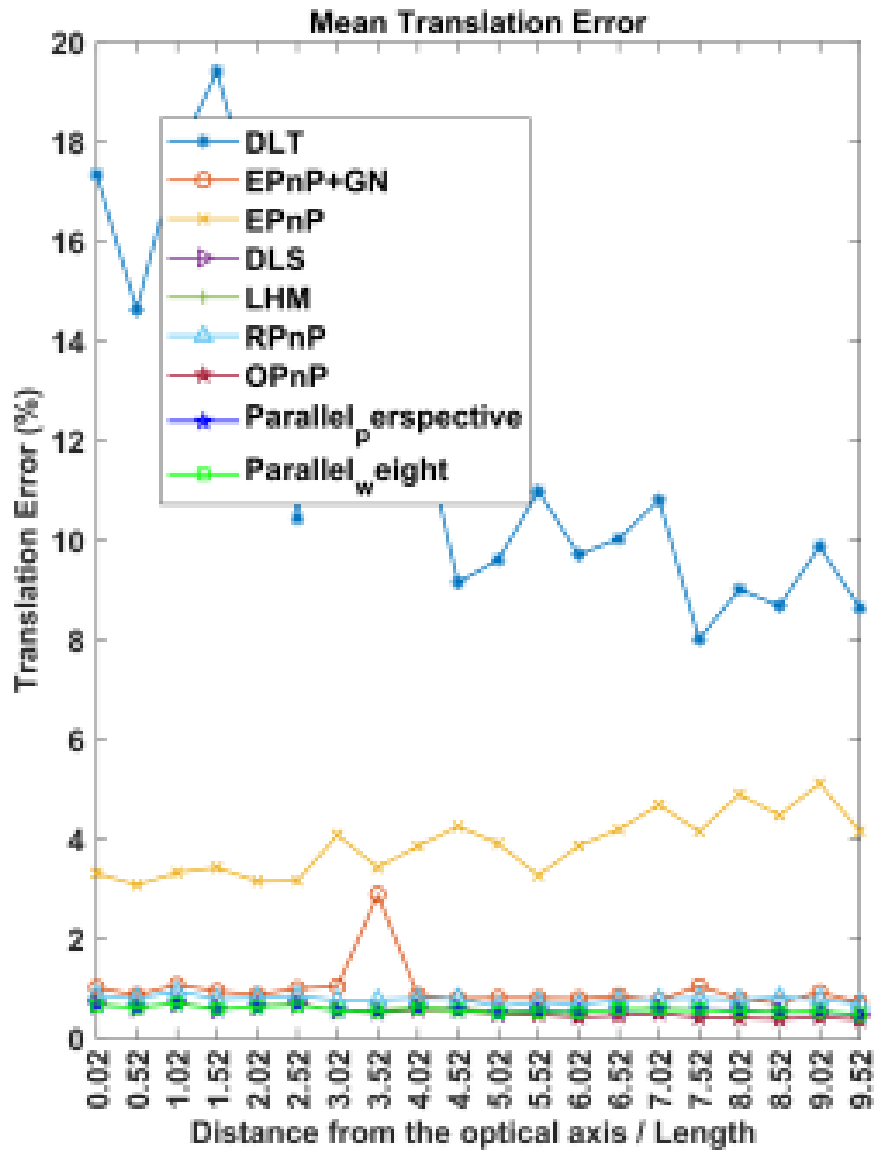

(c)

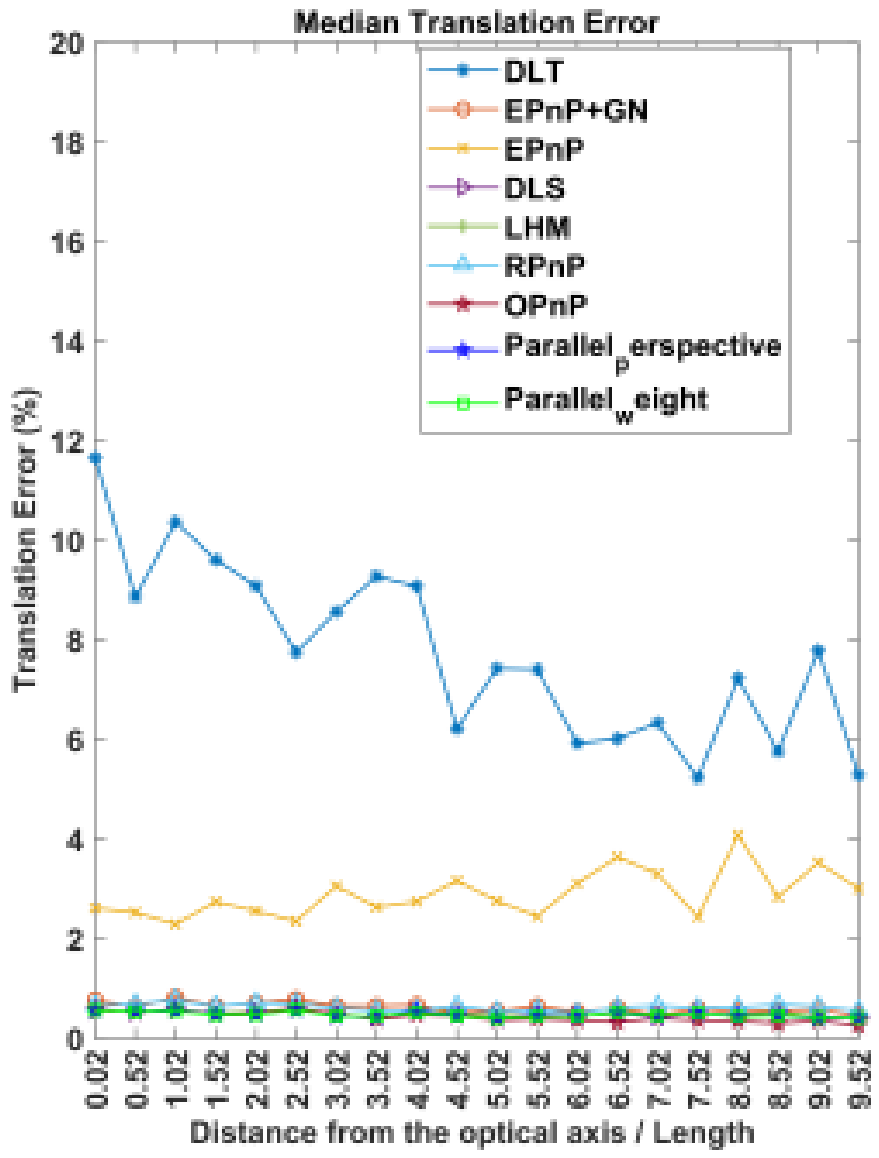

(d)

**Fig. S1.** Effect of the distance from the origin of the coordinate system containing the 6 feature points to the optical axis of the camera on the algorithms (the origin of the coordinate system of the feature points is 200 mm away from the focal point).

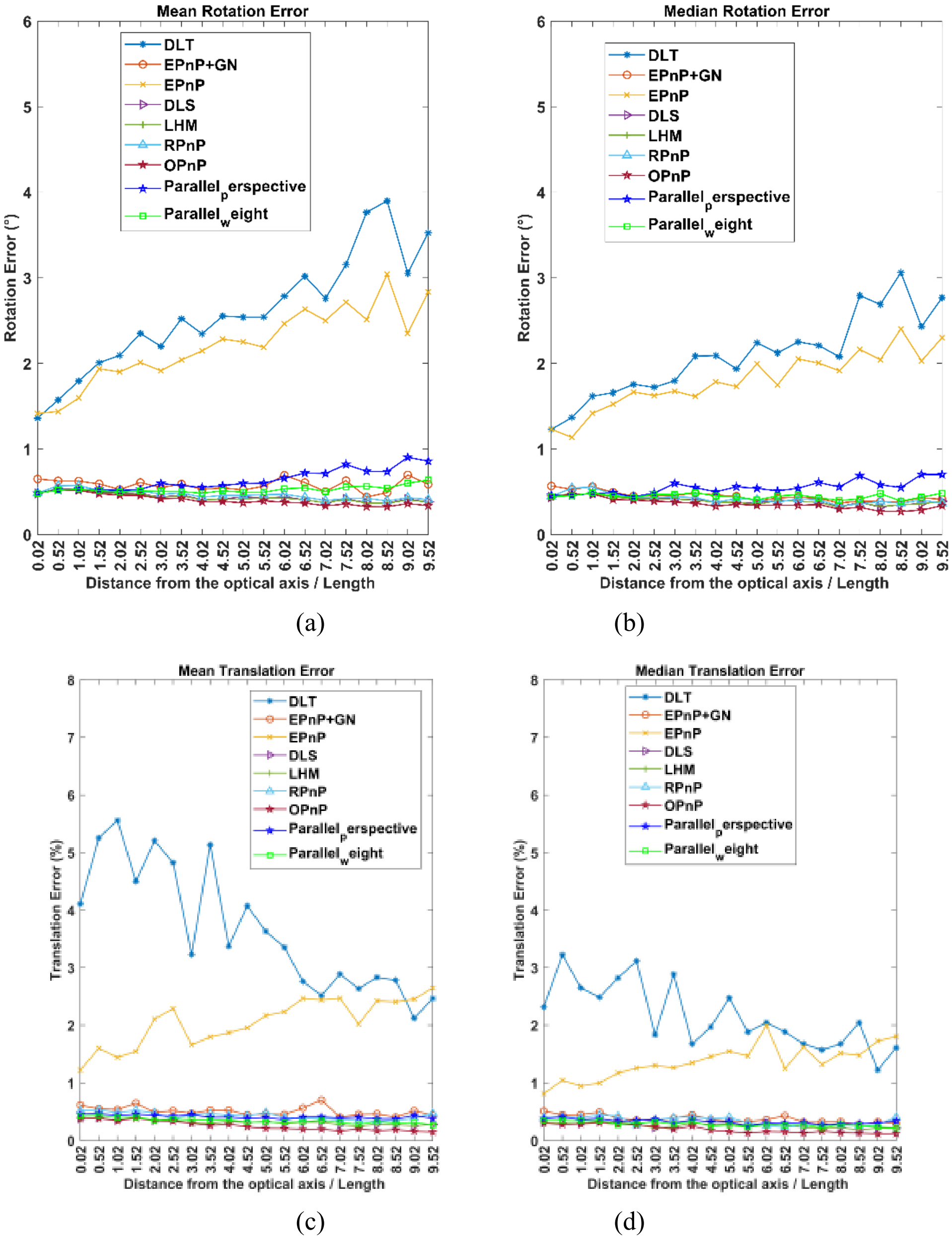

**Fig. S2.** Influence of the distance from the origin of the coordinate system containing the 6 feature points (close to the focus) to the optical axis of the camera (the origin of the coordinate system of the feature points is 100 mm away from the focus) on the algorithms.

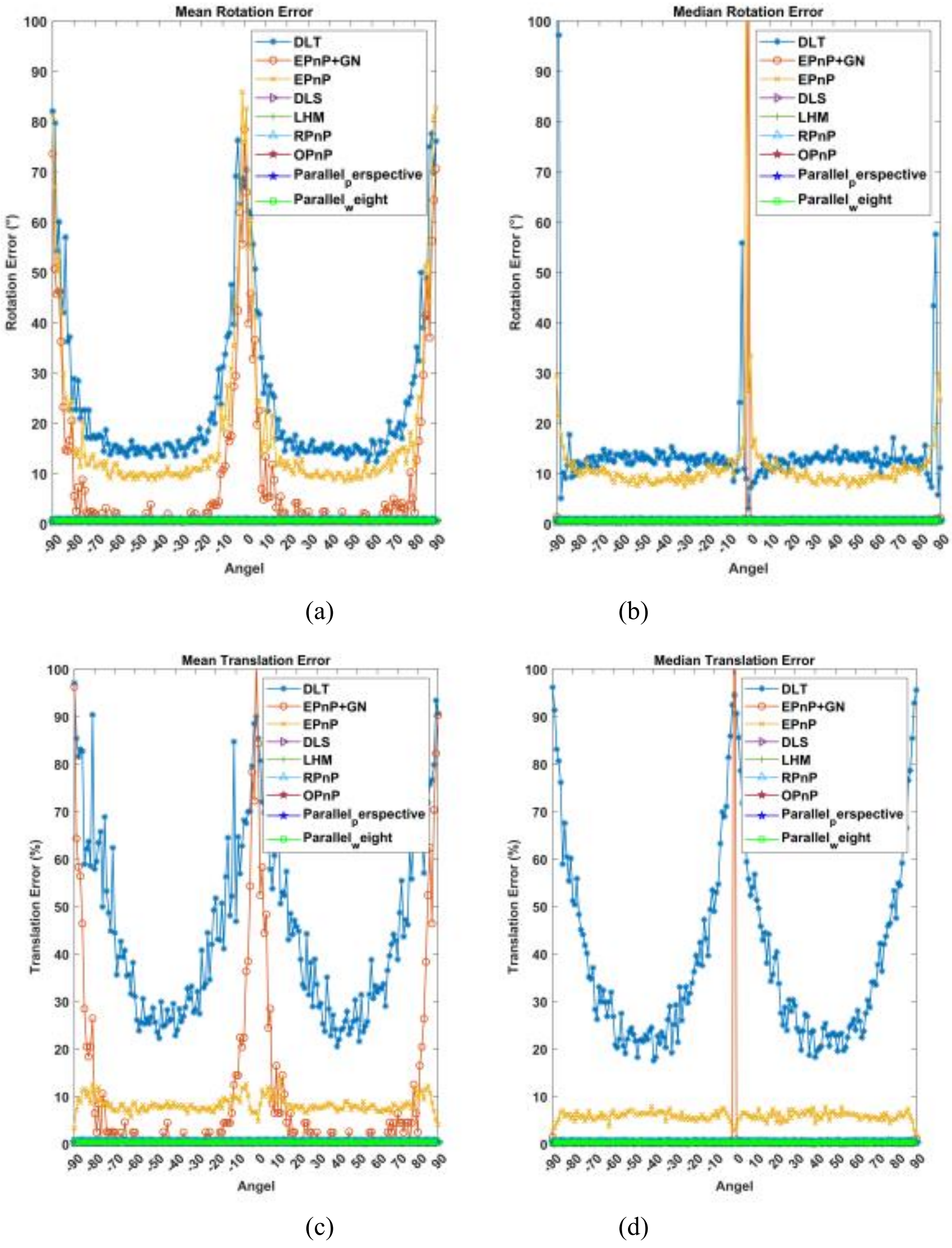

**Fig. S3.** Effects of the angle between the optical axis of the camera and the Z-axis of the coordinate system of the six feature points on the algorithms.

TABLE I
THE MARKER POINT COORDINATE SYSTEM IS 200 MM FROM THE CAMERA

| | Err | DLT | EPnP | EPnP+GN | DLS | LHM | RPnP | OPnP | Parallel | Parallel weight |
|---|---|---|---|---|---|---|---|---|---|---|
| 0 | RE (°) | 2.84 | 2.98 | 1.40 | 1.35 | 1.39 | 1.41 | 1.40 | 1.38 | 1.37 |

TABLE II
THE MARKER POINT COORDINATE SYSTEM IS 400 MM FROM THE CAMERA

| | Err | DLT | EPnP | EPnP+GN | DLS | LHM | RPnP | OPnP | Parallel | Parallel weight |
|---|---|---|---|---|---|---|---|---|---|---|
| 0 | RE(°) | 12.38 | 4.37 | 1.60 | 1.56 | 1.58 | 1.59 | 1.58 | 1.56 | 1.56 |
| | TE (%) | 17.67 | 2.79 | 0.78 | 0.67 | 0.67 | 0.68 | 0.66 | 0.66 | 0.67 |
| 50 | RE(°) | 8.49 | 9.23 | 5.36 | 1.52 | 1.54 | 1.60 | 1.54 | 1.56 | 1.55 |
| | TE (%) | 15.81 | 5.02 | 5.59 | 1.00 | 1.00 | 0.94 | 1.02 | 0.99 | 0.99 |
| 100 | RE(°) | 4.29 | 4.42 | 1.67 | 1.51 | 1.57 | 1.54 | 1.57 | 1.55 | 1.55 |
| | TE (%) | 12.34 | 4.93 | 0.73 | 0.51 | 0.51 | 0.55 | 0.53 | 0.50 | 0.50 |

TABLE III
THE MARKER POINT COORDINATE SYSTEM IS 600 MM FROM THE CAMERA

| | Err | DLT | EPnP | EPnP+GN | DLS | LHM | RPnP | OPnP | Parallel | Parallel weight |
|---|---|---|---|---|---|---|---|---|---|---|
| 0 | RE(°) | 12.44 | 16.40 | 10.83 | 2.98 | 2.99 | 3.00 | 2.99 | 2.97 | 2.98 |
| | TE (%) | 36.59 | 6.33 | 9.92 | 0.72 | 0.72 | 0.76 | 0.79 | 0.78 | 0.77 |
| 50 | RE(°) | 18.36 | 10.90 | 4.77 | 2.93 | 2.88 | 2.91 | 2.88 | 2.87 | 2.87 |
| | TE (%) | 26.96 | 2.84 | 5.81 | 0.50 | 0.51 | 0.78 | 0.56 | 0.54 | 0.54 |
| 100 | RE(°) | 9.35 | 9.10 | 2.88 | 2.90 | 2.91 | 2.95 | 2.91 | 2.92 | 2.92 |
| | TE (%) | 17.47 | 0.60 | 5.33 | 0.52 | 0.52 | 0.63 | 0.54 | 0.52 | 0.52 |

TABLE IV
THE MARKER POINT COORDINATE SYSTEM IS 800 MM FROM THE CAMERA

| | Err | DLT | EPnP | EPnP+GN | DLS | LHM | RPnP | OPnP | Parallel | Parallel weight |
|---|---|---|---|---|---|---|---|---|---|---|
| 0 | RE(°) | 18.83 | 13.06 | 8.62 | 2.93 | 2.83 | 2.81 | 2.83 | 2.82 | 2.81 |
| | TE (%) | 30.05 | 7.44 | 5.98 | 0.47 | 0.47 | 0.70 | 0.48 | 0.46 | 0.43 |
| 50 | RE(°) | 10.70 | 11.80 | 2.86 | 2.93 | 2.88 | 2.92 | 2.88 | 2.88 | 2.88 |
| | TE (%) | 21.44 | 6.34 | 0.53 | 0.43 | 0.43 | 0.60 | 0.41 | 0.41 | 0.41 |
| 100 | RE(°) | 12.61 | 13.57 | 6.37 | 2.81 | 2.77 | 2.82 | 2.77 | 2.75 | 2.75 |
| | TE (%) | 24.16 | 5.33 | 7.74 | 0.52 | 0.53 | 0.73 | 0.57 | 0.54 | 0.50 |

TABLE V
THE MARKER POINT COORDINATE SYSTEM IS 1000 MM FROM THE CAMERA

| | Err | DLT | EPnP | EPnP+GN | DLS | LHM | RPnP | OPnP | Parallel | Parallel weight |
|---|---|---|---|---|---|---|---|---|---|---|
| 0 | RE(°) | 33.79 | 26.68 | 14.68 | 3.72 | 3.63 | 3.64 | 3.71 | 3.59 | 3.59 |
| | TE (%) | 41.04 | 12.14 | 8.17 | 0.59 | 0.59 | 0.78 | 0.63 | 0.58 | 0.58 |
| 50 | RE(°) | 28.31 | 28.62 | 20.08 | 3.16 | 3.10 | 3.29 | 3.11 | 3.05 | 3.06 |
| | TE (%) | 39.08 | 8.16 | 16.61 | 0.52 | 0.52 | 1.03 | 0.56 | 0.55 | 0.54 |
| 100 | RE(°) | 22.92 | 20.01 | 8.70 | 3.37 | 3.24 | 3.22 | 3.24 | 3.18 | 3.19 |
| | TE (%) | 38.32 | 10.01 | 7.52 | 0.88 | 0.88 | 1.22 | 0.83 | 0.87 | 0.87 |

TABLE VI
THE MARKER POINT COORDINATE SYSTEM IS 200 MM FROM THE CAMERA

| | Err | DLT | EPnP | EPnP+GN | DLS | LHM | RPnP | OPnP | Parallel | Parallel weight |
|---|---|---|---|---|---|---|---|---|---|---|
| 0 | RE(°) | 3.82 | 3.64 | 1.27 | 1.34 | 1.37 | 1.35 | 1.37 | 1.29 | 1.29 |
| | TE (%) | 4.54 | 1.79 | 1.41 | 1.14 | 1.14 | 1.31 | 1.15 | 1.19 | 1.19 |

TABLE VII
THE MARKER POINT COORDINATE SYSTEM IS 400 MM FROM THE CAMERA

| | Err | DLT | EPnP | EPnP+GN | DLS | LHM | RPnP | OPnP | Parallel | Parallel weight |
|---|---|---|---|---|---|---|---|---|---|---|
| 0 | RE(°) | 10.40 | 8.43 | 8.40 | 1.84 | 1.77 | 1.81 | 1.78 | 1.73 | 1.73 |
| | TE (%) | 26.94 | 4.46 | 6.42 | 1.37 | 1.36 | 1.51 | 1.34 | 1.31 | 1.31 |
| 50 | RE(°) | 5.47 | 6.20 | 1.42 | 1.35 | 1.38 | 1.47 | 1.38 | 1.37 | 1.37 |
| | TE (%) | 12.07 | 3.50 | 0.98 | 0.65 | 0.65 | 0.86 | 0.69 | 0.68 | 0.68 |
| 100 | RE(°) | 6.63 | 6.81 | 1.58 | 1.69 | 1.69 | 1.66 | 1.70 | 1.74 | 1.74 |
| | TE (%) | 16.20 | 3.28 | 1.01 | 0.83 | 0.84 | 1.06 | 0.89 | 0.88 | 0.88 |

TABLE VIII
THE MARKER POINT COORDINATE SYSTEM IS 600 MM FROM THE CAMERA

| | Err | DLT | EPnP | EPnP+GN | DLS | LHM | RPnP | OPnP | Parallel | Parallel weight |
|---|---|---|---|---|---|---|---|---|---|---|
| 0 | RE(°) | 28.07 | 19.01 | 10.65 | 2.77 | 2.75 | 2.74 | 2.74 | 2.74 | 2.74 |
| | TE (%) | 41.82 | 7.72 | 10.36 | 1.24 | 1.24 | 1.55 | 1.19 | 1.19 | 1.20 |
| 50 | RE(°) | 21.79 | 26.48 | 19.69 | 3.26 | 2.91 | 2.95 | 2.91 | 2.89 | 2.90 |
| | TE (%) | 37.30 | 7.78 | 21.11 | 0.96 | 0.87 | 1.11 | 0.91 | 0.91 | 0.89 |
| 100 | RE(°) | 22.21 | 16.31 | 5.55 | 2.05 | 2.06 | 2.12 | 2.06 | 2.04 | 2.05 |
| | TE (%) | 24.00 | 6.71 | 5.39 | 0.63 | 0.63 | 0.95 | 0.66 | 0.65 | 0.65 |

TABLE IX
THE MARKER POINT COORDINATE SYSTEM IS 800 MM FROM THE CAMERA

| | Err | DLT | EPnP | EPnP+GN | DLS | LHM | RPnP | OPnP | Parallel | Parallel weight |
|---|---|---|---|---|---|---|---|---|---|---|
| 0 | RE(°) | 28.45 | 26.68 | 19.84 | 2.89 | 2.79 | 2.96 | 2.85 | 2.76 | 2.75 |
| | TE (%) | 40.22 | 7.35 | 19.79 | 1.11 | 1.11 | 1.54 | 1.21 | 1.11 | 1.11 |
| 50 | RE(°) | 24.47 | 16.36 | 4.95 | 3.22 | 3.11 | 3.74 | 3.11 | 3.06 | 3.06 |
| | TE (%) | 36.32 | 8.67 | 3.84 | 1.21 | 1.21 | 1.31 | 1.22 | 1.20 | 1.20 |
| 100 | RE(°) | 19.64 | 16.28 | 5.47 | 1.82 | 1.90 | 2.03 | 4.08 | 1.90 | 1.90 |
| | TE (%) | 30.00 | 9.42 | 5.59 | 0.93 | 0.93 | 1.07 | 68.01 | 0.91 | 0.91 |

TABLE X
THE MARKER POINT COORDINATE SYSTEM IS 1000 MM FROM THE CAMERA

| | Err | DLT | EPnP | EPnP+GN | DLS | LHM | RPnP | OPnP | Parallel | Parallel weight |
|---|---|---|---|---|---|---|---|---|---|---|
| 0 | RE(°) | 34.37 | 31.11 | 29.14 | 3.41 | 3.19 | 3.24 | 3.20 | 3.19 | 3.19 |
| | TE (%) | 45.75 | 8.56 | 30.32 | 0.93 | 0.93 | 1.25 | 0.98 | 0.98 | 0.96 |
| 50 | RE(°) | 41.23 | 33.20 | 23.43 | 3.64 | 3.43 | 3.52 | 3.44 | 3.34 | 3.34 |
| | TE (%) | 49.63 | 12.88 | 23.53 | 1.21 | 1.23 | 2.05 | 1.16 | 1.16 | 1.16 |
| 100 | RE(°) | 41.57 | 41.05 | 30.76 | 2.78 | 2.53 | 2.70 | 2.53 | 2.54 | 2.53 |
| | TE (%) | 44. 07 | 15. 39 | 33.57 | 2.12 | 2.12 | 2.50 | 2.35 | 2.33 | 2.32 |

TABLE XI

THE MARKER POINT COORDINATE SYSTEM IS 200 MM FROM THE CAMERA

| | Err | DLT | EPnP | EPnP+GN | DLS | LHM | RPnP | OPnP | Parallel | Parallel weight |
|---|---|---|---|---|---|---|---|---|---|---|
| 0 | RE(°) | 2.96 | 2.78 | 1.81 | 1.78 | 1.64 | 1.78 | 1.61 | 1.55 | 1.55 |
| | TE (%) | 32.26 | 5.99 | 3.05 | 1.70 | 1.70 | 1.91 | 1.56 | 1.25 | 1.25 |

TABLE XII
THE MARKER POINT COORDINATE SYSTEM IS 400 MM FROM THE CAMERA

| | Err | DLT | EPnP | EPnP+GN | DLS | LHM | RPnP | OPnP | Parallel | Parallel weight |
|---|---|---|---|---|---|---|---|---|---|---|
| 0 | RE(°) | 5.75 | 5.14 | 3.55 | 2.07 | 2.03 | 2.03 | 1.99 | 1.97 | 1.97 |
| | TE (%) | 31.47 | 7.49 | 3.60 | 0.74 | 0.74 | 0.96 | 0.74 | 0.75 | 0.75 |
| 50 | RE(°) | 5.56 | 5.32 | 1.66 | 1.61 | 1.54 | 1.91 | 1.52 | 1.52 | 1.52 |
| | TE (%) | 32.57 | 7.28 | 1.51 | 0.88 | 0.88 | 0.99 | 0.84 | 0.76 | 0.76 |
| 100 | RE(°) | 7.15 | 6.56 | 2.10 | 1.80 | 1.72 | 1.95 | 1.72 | 1.73 | 1.73 |
| | TE (%) | 30.26 | 6.26 | 1.88 | 1.37 | 1.37 | 1.40 | 1.28 | 1.17 | 1.17 |

TABLE XIII
THE MARKER POINT COORDINATE SYSTEM IS 600 MM FROM THE CAMERA

| | Err | DLT | EPnP | EPnP+GN | DLS | LHM | RPnP | OPnP | Parallel | Parallel weight |
|---|---|---|---|---|---|---|---|---|---|---|
| 0 | RE(°) | 14.86 | 14.30 | 2.31 | 2.25 | 2.19 | 2.32 | 2.16 | 2.21 | 2.21 |
| | TE (%) | 18.77 | 5.58 | 1.14 | 0.98 | 0.97 | 0.77 | 0.85 | 0.78 | 0.78 |
| 50 | RE(°) | 10.00 | 9.42 | 4.21 | 2.23 | 2.13 | 2.17 | 2.12 | 2.10 | 2.10 |
| | TE (%) | 25.95 | 6.75 | 3.69 | 1.23 | 1.22 | 1.21 | 1.14 | 1.06 | 1.06 |
| 100 | RE(°) | 7.94 | 6.51 | 1.95 | 3.15 | 1.72 | 1.70 | 1.73 | 1.76 | 1.76 |
| | TE (%) | 29.72 | 7.04 | 1.58 | 1.88 | 1.01 | 1.12 | 1.02 | 1.01 | 1.01 |

TABLE XIV
THE MARKER POINT COORDINATE SYSTEM IS 800 MM FROM THE CAMERA

| | Err | DLT | EPnP | EPnP+GN | DLS | LHM | RPnP | OPnP | Parallel | Parallel weight |
|---|---|---|---|---|---|---|---|---|---|---|
| 0 | RE(°) | 11.22 | 10.63 | 3.80 | 2.20 | 2.11 | 2.18 | 2.10 | 2.13 | 2.13 |
| | TE (%) | 23.81 | 7.72 | 2.99 | 0.64 | 0.64 | 0.62 | 0.60 | 0.57 | 0.57 |
| 50 | RE(°) | 12.44 | 10.36 | 2.47 | 2.24 | 2.20 | 2.27 | 2.24 | 2.20 | 2.20 |
| | TE (%) | 23.37 | 7.36 | 1.22 | 0.86 | 0.87 | 1.14 | 0.91 | 0.89 | 0.89 |
| 100 | RE(°) | 12.63 | 12.86 | 3.96 | 2.01 | 2.01 | 2.06 | 3.96 | 2.06 | 2.06 |
| | TE (%) | 21.52 | 6.48 | 3.45 | 0.92 | 0.92 | 1.24 | 214.86 | 0.98 | 0.98 |

TABLE XV
THE MARKER POINT COORDINATE SYSTEM IS 1000 MM FROM THE CAMERA

| | Err | DLT | EPnP | EPnP+GN | DLS | LHM | RPnP | OPnP | Parallel | Parallel weight |
|---|---|---|---|---|---|---|---|---|---|---|
| 0 | RE(°) | 21.66 | 18.54 | 5.86 | 2.62 | 2.66 | 2.72 | 2.66 | 2.67 | 2.67 |
| | TE (%) | 32.56 | 7.37 | 6.26 | 1.57 | 1.56 | 1.99 | 1.54 | 1.63 | 1.63 |
| 50 | RE(°) | 13.91 | 13.91 | 4.04 | 2.24 | 2.13 | 2.20 | 2.13 | 2.10 | 2.10 |
| | TE (%) | 22.37 | 7.27 | 3.10 | 0.69 | 0.67 | 0.76 | 0.68 | 0.67 | 0.67 |
| 100 | RE(°) | 20.66 | 19.68 | 7.16 | 5.17 | 1.88 | 2.01 | 1.88 | 2.00 | 2.00 |